\documentclass{article}  
\usepackage{iclr2024_conference,times}
\usepackage{amsmath,amsfonts,bm}

\def\eqref#1{equation~\ref{#1}}
\def\1{\bm{1}}

\DeclareMathAlphabet{\mathsfit}{\encodingdefault}{\sfdefault}{m}{sl}
\SetMathAlphabet{\mathsfit}{bold}{\encodingdefault}{\sfdefault}{bx}{n}

\newcommand{\R}{\mathbb{R}}

\newcommand{\KL}{D_{\mathrm{KL}}}

\usepackage{graphicx}
\usepackage{caption}
\usepackage{subcaption}%
\usepackage[pdfusetitle]{hyperref}
\usepackage{xcolor}
\hypersetup{
    colorlinks,
    linkcolor={red!50!black},
    citecolor={blue!50!black},
    urlcolor={blue!80!black}
}
\usepackage{url}
\usepackage{cleveref}
\usepackage{courier}
\usepackage{listings}
\usepackage{booktabs}
\usepackage{xcolor}

\renewcommand{\P}{\mathbb{P}}
\newcommand{\LD}{\text{LD}}
\newcommand{\pt}{\text{pt}}
\newcommand{\cl}{\text{cl}}
\newcommand{\xco}{X_\text{corrupt}}
\newcommand{\xcl}{X_\text{clean}}

\usepackage{inconsolata}
\usepackage{url}

\usepackage{footnote}
\usepackage{footmisc}
\makeatother
\renewcommand{\headrulewidth}{0pt}

\makeatletter
\def\url@leostyle{%
  \@ifundefined{selectfont}{\def\UrlFont{\sf}}{\def\UrlFont{\small\ttfamily}}}
\makeatother
\usepackage[shortlabels]{enumitem}
\title{Towards Best Practices of  Activation Patching  in Language Models:  Metrics and Methods}

\iclrfinalcopy 
\author{Fred Zhang\thanks{Work done while interning at Google.}\\
UC Berkeley\\
\texttt{z0@berkeley.edu} \\
\And
Neel Nanda \\
Independent\\
\texttt{neelnanda27@gmail.com}
}

\begin{document}

\maketitle

\begin{abstract}
Mechanistic interpretability seeks to understand the  internal mechanisms of machine learning models, where  localization---identifying the important model components---is a key step. Activation patching, also known as causal tracing or interchange intervention, is a standard technique  for this task \citep{vig2020investigating}, but the literature contains many variants    with little consensus on the choice of hyperparameters or methodology. In this work, we systematically examine the impact of   methodological details in activation patching, including evaluation metrics and corruption methods. In several settings of localization and circuit discovery in language models, we find that   varying these hyperparameters could lead to disparate interpretability results. 
 Backed by  empirical observations, we     give conceptual arguments for why certain metrics or methods may be preferred. 
 Finally,  we provide recommendations for the best practices of activation patching going forwards.
\end{abstract}

\section{Introduction}
% Transformer-based language models
% \citep{vaswani2017attention, brown2020language} have demonstrated remarkable   capabilities and yet largely remain black boxes. 
% Their deployment in high-stake settings, therefore, calls for research in explaining their behaviors.
Mechanistic interpretability (MI)   aims to  unravel 
complex machine learning models   by reverse engineering  their internal mechanisms down to human-understandable algorithms \citep{geiger2021causal, AnthropicMechanisticEssay, wang2022interpretability}.  
With such    understanding, we can better  identify and fix model errors \citep{vig2020investigating,hernandez2021natural, meng2022locating, hase2023does}, steer model outputs \citep{li2023inference} and explain   emergent behaviors \citep{nanda2022progress,barak2022hidden}.

A basic goal in MI is {localization}: identify the  specific model components responsible for particular functions.
Activation patching, also known as causal tracing, interchange intervention, causal mediation analysis or representation denoising,    is a standard tool for localization in   language models  \citep{vig2020investigating,meng2022locating}. The method attempts to pinpoint activations that causally affect on the output. Specifically, it involves 3 forward passes of the model: (1) on a clean prompt while  caching the  latent activations; (2) on a corrupted prompt; and (3) on the corrupted prompt but replacing the  activation of a specific model component by its clean cache.  For instance,  the clean prompt can be  ``{The Eiffel Tower is in}'' and the corrupted one with the subject   replaced by ``The Colosseum''. If the model outputs ``Paris'' in step (3) but not in (2),  then it suggests that the specific component being patched is   important for producing  the answer \citep{vig2020investigating, pearl2001direct}.  

This technique has been widely applied for language model interpretability. For example, \cite{meng2022locating,geva2023dissecting} seek to understand which model weights store and process factual information. 
\cite{wang2022interpretability, hanna2023does, lieberum2023does}   perform circuit analysis: identify the sub-network within   a model's computation graph that implements a specified behavior. All these works leverage activation patching or its variants as a foundational technique.

Despite  its broad applications across the literature, there is little consensus on the methodological details of activation patching.
In particular, each paper tends to use its own  method of generating corrupted prompts  and the metric of evaluating patching effects.  Concerningly, this lack of standardization leaves open the possibility that prior interpretability results may be highly sensitive to the hyperparameters they adopt. 
In this work, we study   the impact of varying the metrics and methods in activation patching, as a step towards understanding   best practices. To our knowledge, this is the first such systematic study of the technique.
 
Specifically, we identify three degrees of freedom in activation patching. First, we focus on   the approach of generating   corrupted prompts and evaluate two prominent methods from the literature:
\begin{itemize}[leftmargin=1em]
\item Gaussian noising (GN) adds
a large Gaussian noise to the token embeddings
of the   tokens that contain the key information   to 
completing a prompt, such as its subject \citep{meng2022locating}.
    \item Symmetric token replacement (STR)  swaps these key tokens with semantically related ones; 
    for example, ``{The Eiffel Tower}''$\rightarrow$``{The Colosseum}'' \citep{vig2020investigating, wang2022interpretability}.
\end{itemize}
Second, we examine  the choice of   metrics for measuring    the effect of patching and compare   probability and logit difference; both have found applications in the literature \citep{meng2022locating,wang2022interpretability, conmy2023towards}. Third, we study   sliding window patching, which jointly restores the activations of  multiple MLP layers, a technique used by \cite{meng2022locating,geva2023dissecting}. 

We empirically examine the impact of these   hyperparameters on several   interpretability tasks, including   factual recall     \citep{meng2022locating} and  circuit discovery for   indirect object identification (IOI)   \citep{wang2022interpretability}, greater-than \citep{hanna2023does}, Python docstring  completion  \citep{docstring} and basic arithmetic \citep{stolfo2023understanding}. In each setting, we apply   methods  distinct from the original studies and assess  how different interpretability results arise from these variations.

\paragraph{Findings}
Our contributions uncover nuanced discrepancies within activation patching techniques applied to language models. 
On corruption method, we show that GN and STR can lead to inconsistent localization and circuit discovery outcomes (\autoref{sec:corruption-diff}). 
Towards explaining the gaps, we posit that GN breaks model's internal mechanisms by putting it off distribution.
We give tentative evidence for this claim in the setting of IOI circuit discovery (\autoref{sec:ood}). We believe that this is a fundamental concern in using GN corruption for activation patching. 
On evaluation metrics, we provide an analogous set of differences between logit difference and probability (\autoref{sec:metrics}), including an observation that    probability can overlook negative model components that hurt performance.

Finally, we  compare sliding window patching with patching individual layers and summing up their effects. We find    the sliding window method     produces more pronounced  localization   than single-layer patching and discuss the conceptual differences between these two approaches (\autoref{sec:sliding}).

\paragraph{Recommendations for practice} At a high-level, our findings highlight the sensitivity of activation patching to methodological details. 
Backed by our analysis, we make several recommendations on the application of activation patching in language model interpretability (\autoref{sec:discussion}). 
We advocate for STR, as it supplies  in-distribution corrupted prompts that  help to preserve consistent model behavior. 
On evaluation metric, we recommend logit difference, as we argue that it offers fine-grained control over the localization outcomes and is capable of detecting negative modules.
\section{Background}\label{sec:background}
\begin{figure}[t!]
\vspace{-3em}
    \centering
    \begin{subfigure}[t]{0.55\linewidth}
        \includegraphics[width=\linewidth]{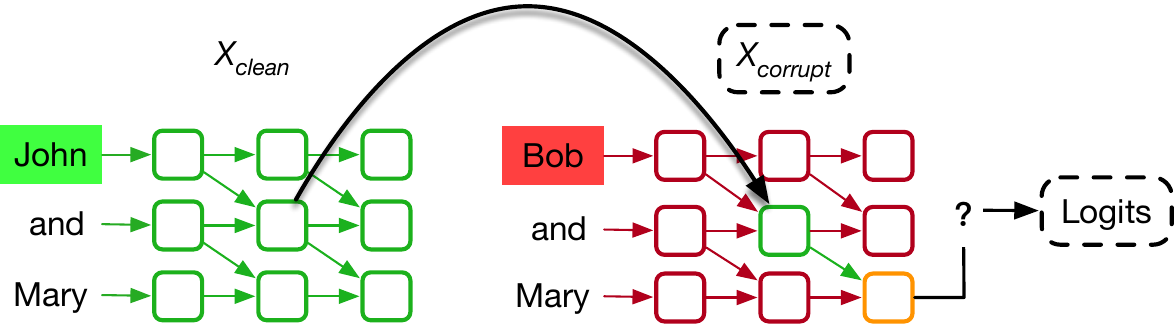}
  \caption{Activation patching intervenes on latent states}
    \label{fig:patched}
    \end{subfigure}
    \qquad
    \begin{subfigure}[t]{0.3\linewidth}
    \hspace{0.7em}
  \includegraphics[width=0.75\linewidth]{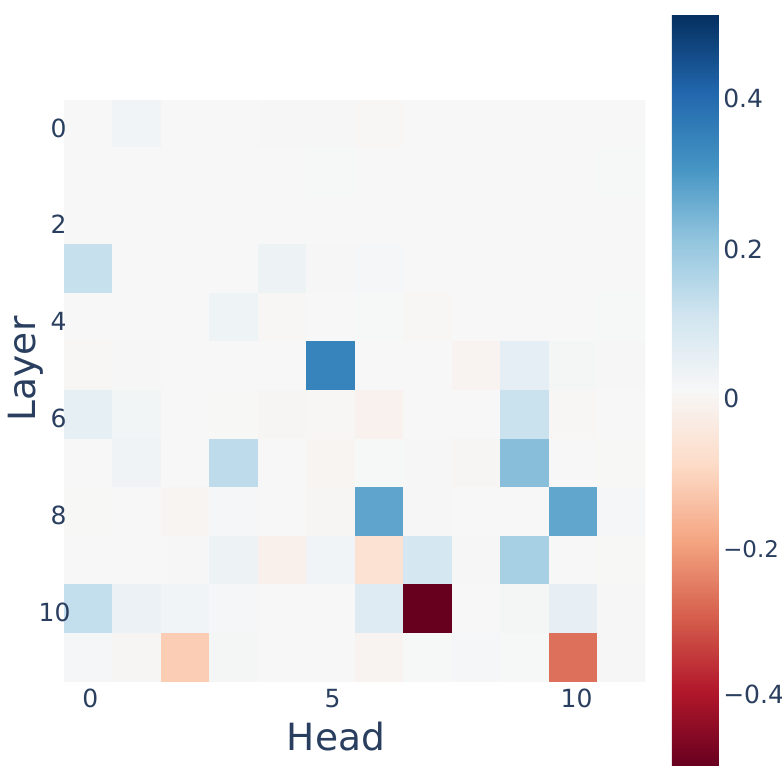}
  \caption{Patching attention heads}
      \label{fig:example-plot}

    \end{subfigure}
    \caption{\textbf{The workflow of activation patching} for localization: run the  intervention procedure (a) on every relevant component, such as all the attention heads, and plot the effects (b).}
\end{figure}
% In this section, we   introduce activation patching   and define the problem settings  we work with.  

\subsection{Activation patching} \label{sec:background-patching}
Activation patching identifies the important model components by intervening on their latent  activations. The method        involves a clean prompt ($X_\text{clean}$, e.g.,``The Eiffel Tower is in'') with an associated answer   $r$ (``Paris''), a {corrupted prompt} ($X_\text{corrupt}$, e.g., ``The Colosseum is in''), and three model runs:
\begin{enumerate}[(1), leftmargin=2em]
    \item {Clean run}: run the model on $X_\text{clean}$ and cache activations of a set of given model components, such as MLP or attention heads outputs.
    \item {Corrupted run}: run the model on  $X_\text{corrupt}$ and record the model outputs.
    \item {Patched run}:  run the model on $X_\text{corrupt}$ with a specific model component's activation restored from the cached value of the clean run (\autoref{fig:patched}).
\end{enumerate}
Finally, we evaluate the patching effect, such as $\P$(``Paris'') in the patched run (3) compared to the corrupted run (2). 
Intuitively, corruption hurts model performance while patching restores it. Patching effect  measures
how much the  patching intervention restores performance, which indicates the importance of the activation. 
We can iterate this procedure over a collection of components (e.g., all attention heads), resulting in a plot that highlights the   important ones (\autoref{fig:example-plot}).
 
\paragraph{Corruption methods} To generate $\xco$, GN adds Gaussian noise $\mathcal N(0,\nu)$ to the   embeddings of certain key tokens, where $\nu$ is $3$ times the standard deviation of the token
embeddings from the textset.  STR replaces the key  tokens by similar ones with equal sequence length.   In   STR,   let $r'$  denote the answer  of $X_\text{corrupt}$ (``Rome'').
    All implementations of STR in this paper yield in-distribution prompts such that $\xco$ is identically distributed as a fresh draw of a clean prompt.
 \paragraph{Metrics} The patching effect  is defined as the gap  of the model performance between the corrupted and patched run, under an evaluation metric. Let $\cl$, $*$, $\pt$ be the clean, corrupted and patched run. 
\begin{itemize}[leftmargin=1em]
\item 
 Probability: $\P(r)$; e.g., $\P(\text{``Paris''})$. The patching effect is $\P_{\text{pt}} (r) - \P_{*}(r)$; 
 \item 
Logit difference: $\LD(r,r')=\text{Logit}(r) - \text{Logit}(r')$; e.g., $\text{Logit}(\text{``Paris''}) - \text{Logit}(\text{``Rome''})$. 

 The patching effect is given by $\LD_\text{pt}(r,r') - \LD_*(r,r')$.  
 Following \cite{wang2022interpretability}, we always normalize this by $\LD_\cl(r,r') - \LD_{*}(r,r')$,
 so it typically lies in $[0,1]$, where $1$ corresponds to fully restored performance and $0$ to  the corrupted run performance.
\item KL divergence: $\KL(P_{\text{cl}} || P)$, the Kullback-Leibler (KL) divergence from the probability distribution of model outputs in the clean run. The patching effect is $\KL(P_{\text{cl}} || P_*) -\KL(P_{\text{cl}} || P_{\text{pt}})$.
\end{itemize}
GN does not provide a corrupted prompt with a well-defined answer $r'$ (``Rome''). To make a fair comparison, the same $r'$ is used for evaluating the logit difference metric under GN. 

\subsection{Problem settings}
\paragraph{Factual recall}  In  the setting of factual association, the model is  prompted to fill in factual information, e.g., ``The Eiffel Tower is   in''.   
\cite{meng2022locating}
  posits that Transformer-based language models complete factual recall (i) at middle MLP layers and (ii) specifically at the processing of the subject’s last token. In this work, we do not treat the hypothesis as ground-truth but rather reevaluate it using other approaches than what was attempted by \cite{meng2022locating}.

\paragraph{IOI} An IOI sentence  involves  an initial dependent
clause, e.g., ``When John and Mary went to the office'', followed by a main clause, e.g., ``John gave a book to Mary.'' 
In this case, the indirect object (IO) is ``Mary'' and the subject (S) ``John''.   
The IOI task is to predict the final token in the sentence to be the
IO.  
We use S1 and S2 to refer to the first and second occurrences of the subject (S).

We let $p_{\text{IOI}}$   denote the distribution of IOI sentences of \cite{wang2022interpretability} containing single-token names.
GPT-2 small performs well on  $p_{\text{IOI}}$ and \cite{wang2022interpretability}
discovers a circuit within the model for this task. The circuit consists of attention heads. This is also the focus of our experiments, where
we uncover nuanced differences when using different techniques to  replicate  their result.

% In addition, an overview of the transformer architecture is given in \autoref{sec:arc}.
 % \paragraph{Other tasks} We also validate our high-level findings on the Python docstring circuit in a $4$-layer attention-only model \citep{docstring} and in  a setting of localizing arithmetic   capabilities in GPT-J \citep{stolfo2023understanding}. The details  are in   \autoref{sec:arithmetic} and \autoref{sec:docstring}.

\section{Corruption methods}
\label{sec:corruption}
In this section, we evaluate GN and STR on     localizing factual recall in GPT-2 XL  and discovering the  IOI circuit in GPT-2 small. 
\paragraph{Experiment setup}
For factual recall, we investigate \cite{meng2022locating}'s hypothesis 
that model computation is concentrated at early-middle MLP layers (by processing the last subject token). 
Specifically, we corrupt the subject token(s) to generate $\xco$. In the patched run, we override the MLP activations at the last subject token.  Following \cite{meng2022locating, hase2023does}, at each layer we restore a set of $5$ adjacent MLP layers. (More results on other window sizes can be found in \autoref{sec:b1}. We   examine sliding window patching more closely in \autoref{sec:sliding}.)

For IOI circuit discovery, we follow \cite{wang2022interpretability} and focus on the role of attention heads. Corruption is applied to the S2 token.
Then we patch a single attention head's output (at all positions) and iterate over all heads in this way. 
To avoid relying  on visual inspection, we say that a head is \textit{detected} if its patching effect is $2$ standard deviation (SD) away from the mean effect. 

\paragraph{Dataset and corruption method}
STR requires pairs of $\xcl$ and $\xco$ that are semantically similar. To perform STR,  we construct   \textsc{PairedFacts} of $145$   pairs of prompts  on factual recall.   All the prompts are in-distribution, as they are selected from the original dataset of \cite{meng2022locating}; see \autoref{sec:exp-details} for details. GPT-2 XL achieves an average of $49.0\%$ accuracy on this dataset.

For the IOI circuit, we use the   $p_{\text{IOI}}$ distribution to sample the clean prompts.  For STR, we replace S2 by IO to construct $\xco$ such that $\xco$ is still a valid  in-distribution IOI sentence. For GN, we add noise to the S2's token embedding. The experiments are averaged over $500$ prompts.

\subsection{Results on corruption methods}
\label{sec:corruption-diff}
\paragraph{Difference in MLP localization} 
\begin{figure}
\vspace{-3em}
    \centering
    \begin{subfigure}[b]{0.2\linewidth}
        \includegraphics[width=\linewidth]{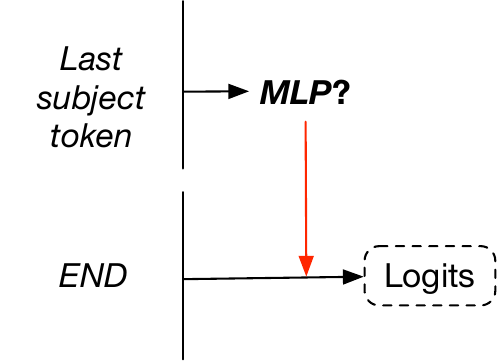}
        \caption{Patching MLP   at the last subject token.}
    \end{subfigure}
    \qquad
    \begin{subfigure}[b]{0.35\linewidth}
    \centering
    \includegraphics[width=\linewidth]{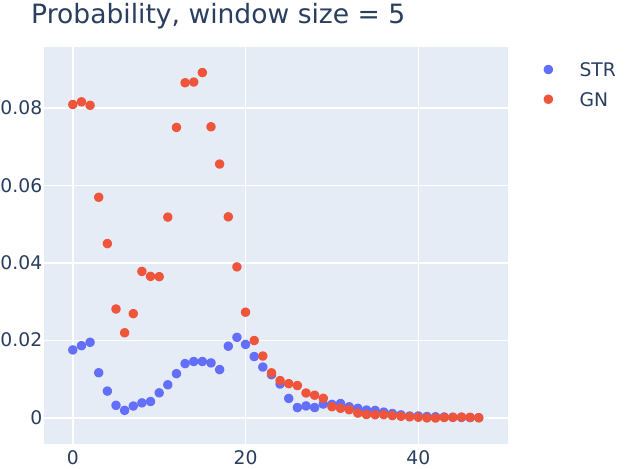}
    \caption{Probability as the metric}
    \end{subfigure}
    \begin{subfigure}[b]{0.35\linewidth}
    \centering
        \includegraphics[width=\linewidth]{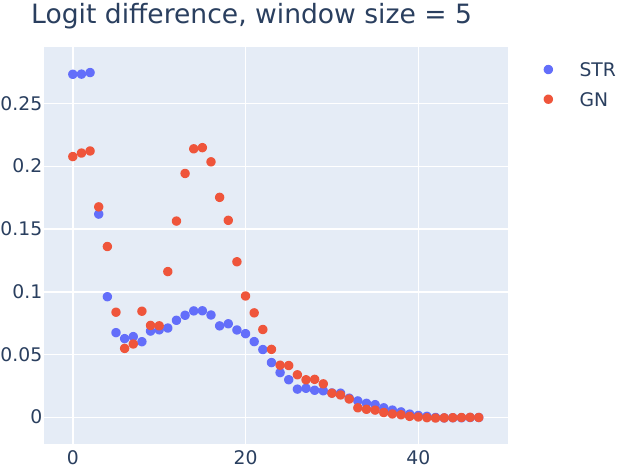}
        \caption{Logit difference as the metric}
    \end{subfigure}
    \caption{\textbf{Disparate MLP patching effects for factual recall in GPT-2 XL}. (a) We patch MLP activations at the last subject token. (b)(c) The patching effects  using different corruption methods with a window size of $5$. {STR  suggests much a weaker peak, regardless of the evaluation metric}.\protect\footnotemark}
    \label{fig:main-mlp-fact}
\end{figure}
\footnotetext{The effects on the first $3$ layers are large simply because  MLP0 has significant influence on the model's outputs in GPT-2, regardless of the task \citep{wang2022interpretability,hase2023does}, so it is not the focus here.}
For patching MLPs in the factual association setting, \cite{meng2022locating}  show that the effects concentrate at  early-middle layers, where they apply GN as the corruption method. Our main finding is that the picture can be largely different by switching the corruption method, regardless of the choice of metric. 
In \autoref{fig:main-mlp-fact}, we plot the patching effects for both metrics. Notice that the clear peak around layer 16 under GN is not salient at all under STR. 

This is a robust phenomenon: across  window sizes, we find the peak value of GN to be 2\texttimes--5\texttimes\ higher than STR; see Appendix \ref{sec:b1} for further plots on GPT-2 XL in this setting. 

These findings  illustrate potential discrepancies between the two  corruption techniques in drawing interpretability conclusions. 
We do not, though, claim that results from GN are   illusory or overly inflated.
In fact,  GN does {not} {always} yield sharper peaks than STR.
For certain basic arithmetic tasks in GPT-J,  STR can   show stronger concentration in patching MLP activations; see \autoref{sec:arithmetic}.

\paragraph{Difference in circuit discovery} 
We focus on discovering the main classes of attention heads in the IOI circuit, including (Negative) Name Mover (NM),  Duplicate Token (DT), S-Inhibition (SI), and Induction Heads.   The results are summarized in \autoref{tab:ioi} and more details  in \autoref{sec:further-ioi}.  
\begin{table}[ht]
\vspace{-2.4em}
\centering
\begin{tabular}{l|l||c|c|c|c|c}
\toprule
\textbf{Corruption} & \textbf{Metric} & \multicolumn{1}{l|}{NM} & \multicolumn{1}{l|}{DT} & SI & \multicolumn{1}{l|}{Negative NM} & \multicolumn{1}{l}{Induction} \\ \midrule
STR & Probability & $\mathbf{1/3}$ & $\mathbf{0/2}$ & $\mathbf{3/4}$ & $\mathbf{1/2}$ & $1/2$ \\
GN $^\dagger$ & Probability & $\mathbf{0/3}$ & $\mathbf{1/2}$ & $\mathbf{2/4}$ & $\mathbf{2/2}$ & $1/2$ \\ \midrule
STR & Logit difference & $1/3$ & $\mathbf{0/2}$ & $3/4$ & $2/2$ & $1/2$ \\
GN & Logit difference & $1/3$ & $\mathbf{1/2}$ & $3/4$ & $2/2$ & $1/2$ \\ \midrule
STR & KL divergence & $\mathbf{1/3}$ & $0/2$ & $\mathbf{3/4}$ & $2/2$ & $1/2$ \\
GN $^\dagger$ & KL divergence & $\mathbf{0/3}$ & $0/2$ & $\mathbf{2/4}$ & $2/2$ & $1/2$ \\
\bottomrule
\end{tabular}
\caption{\textbf{Inconsistency in circuit discovery from activation patching   on the IOI task}. We patch the attention heads outputs and list  the detections of each class.  \footnotesize{$^\dagger$Also detect   0.10, a fuzzy Duplicate Token Head,  as \textit{negatively} influencing model performance. We expect    it to be positive \citep{wang2022interpretability}. }}
\label{tab:ioi}
\end{table}

Most importantly, we observe that  {STR and GN produce inconsistent discovery results}. In particular, for any fixed metric, STR and GN detect   different sets of heads as important,   highlighted in \autoref{tab:ioi}.

We remark that all the detections  are in the IOI circuit as found by \cite{wang2022interpretability}. However, the discovery we achieved here appear far from complete, with some critical misses such as NM. This suggests that the extensive manual inspection and the use of path patching, a more   surgical  patching method, are   both necessary to fully discover   the IOI circuit. 

We also validate our high-level conclusions on the Python docstring  \citep{docstring} and the greater-than \citep{hanna2023does} task. In particular, we  find   GN can produce highly noisy localization outcomes in these settings; see \autoref{sec:docstring} and \autoref{sec:gt} for details.
\subsection{Evidence for   OOD behavior in Gaussian noise corruption}
\label{sec:ood}
We suspect that the gaps between the corruption methods can be  attributed partly to model's OOD behavior under GN corruption. In particular, the Gaussian noise may break model's internal mechanisms by introducing OOD inputs to the layers. We  now  give some tentative evidence for this hypothesis. Following the notation of \cite{wang2022interpretability},  a head is denoted by ``layer.head''.
\paragraph{Negative detection of 0.10 under GN}
Although most localizations we obtain above seem aligned with the findings of \cite{wang2022interpretability},
a major anomaly    in the GN experiment is the ``negative'' detection of 0.10. In particular, probability and KL divergence   suggest that it contributes negatively to  model performance. (Logit difference also assigns a negative  effect, though to a lesser degree; see \autoref{fig:str-010}.) This is not observed at all in the experiments with STR corruption. 

The detection is in the wrong direction, given the    evidence  from \cite{wang2022interpretability}  that 0.10  \textit{helps} with IOI; 
on clean prompts, it is active   at  S2, attends to S1 and signals this duplication. 
However, by visualizing the attention patterns, we find that this effect largely disappears under GN corruption.
We intuit that the Gaussian noise is strongest at influencing early layers, and 0.10's behavior may be broken here, 
since it directly receives the noised  token embeddings from the residual stream.

\begin{figure}[t]
% \vspace{-2em}
    \centering
    \begin{subfigure}{0.27\linewidth}
    \centering
        \includegraphics[width=\linewidth]{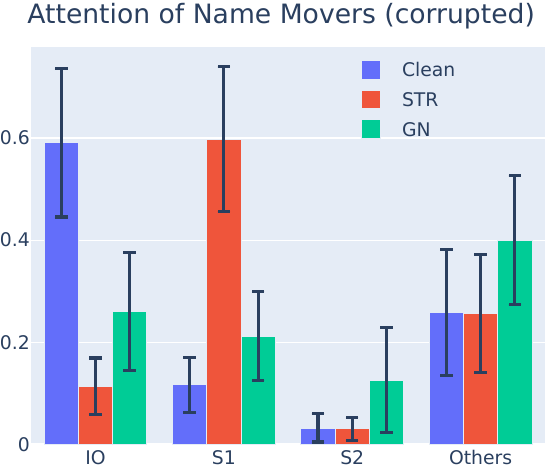}
        \caption{Corrupted run}
        \label{fig:ioi-mh-corrupted}
    \end{subfigure}
    \qquad
      \begin{subfigure}{0.25\linewidth}
    \centering
    \includegraphics[width=\linewidth]{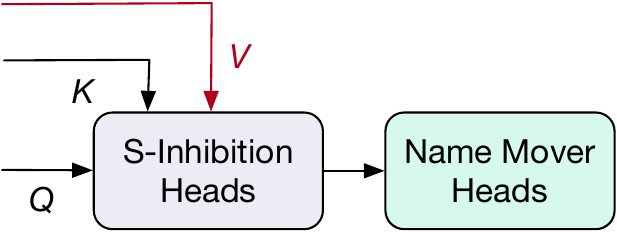}
    \caption{We patch the value matrices of all the SI heads and examine the impact on NMs' attention patterns.}
    \label{fig:patch-nmh}
    \end{subfigure}
    \qquad
     \begin{subfigure}{0.27\linewidth}
    \centering
        \includegraphics[width=\linewidth]{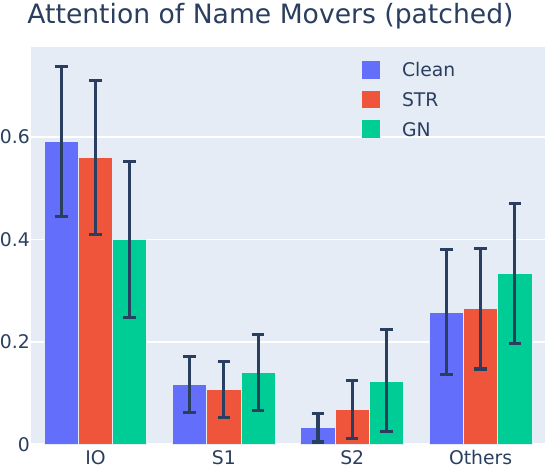}
        \caption{Patched run}
        \label{fig:patched-ioi-nhm}
    \end{subfigure}
    \caption{\textbf{Attention of the Name Movers} from the last token, in corrupted and patched runs. 
    }
    \label{fig:ioi-nmh}
\end{figure}

\paragraph{Attention of   Name Movers}
To exhibit the   OOD behavior of the model internals under GN corruptions, we examine the Name Mover (NM) Heads, a class of attention heads that directly affects the model's logits in the IOI circuit \citep{wang2022interpretability}. 
NMs are active at the last token and copy what they attend to.  We plot the attention of NMs in clean and corrupted runs in \autoref{fig:ioi-mh-corrupted}.

Indeed, on $500$ clean IOI prompts, the NMs assign an average of $0.58$ attention probability to IO. 
In the corrupted runs,  since STR simply exchanges  IO by S1, the attention patterns of NMs are preserved (with the role of IO and S1 switched). On the other hand, with GN corruption, we see that the attention is shared between IO and S1 ($0.26$ and $0.21$).   This suggests that GN not only removes the relevant information but also  disrupts the internal mechanism of NMs on IOI sentences.

To take a deeper dive,  \cite{wang2022interpretability} shows that the output of NMs is determined largely by the values of the S-Inhibition Heads.
Indeed, we  can fully recover model's logit on IO in STR (logit difference: $1.04$) by restoring the values of the S-Inhibition Heads (\autoref{fig:patch-nmh}). 
The same intervention, however,  is   fairly unsuccessful under GN (logit difference: $0.49$).

Towards explaining this gap, we again examine the attention of NMs.
\autoref{fig:patched-ioi-nhm} shows that patching nearly   restores the NMs' in-distribution attention pattern under STR, but fails under GN corruption.  
We speculate that GN introduces further corrupted information flowing into the NMs such that restoring the clean activations of S-Inhibition Heads cannot correct their behaviors.
\section{Evaluation Metrics}
\label{sec:metrics}
We now study the choice of evaluation metrics in activation patching. 
We perform two   experiments that highlight   potential gaps between logit difference and probability. Along the way, we provide a conceptual argument for why probability can overlook negative components in certain settings.

\subsection{Localizing factual recall with logit difference}
\label{sec:fact-last}
The prior work of \cite{meng2022locating}   hypothesizes that factual association is processed  at the last subject token. Motivated by this claim, we extend our previous experiments to patching the MLP outputs at all token positions and consider the effect of changing evaluation metrics.

\paragraph{Experimental setup} We apply the same setting as in \autoref{sec:corruption}. We extend our MLP patching experiments to all token positions and again use logit difference and probability as the metric. 

\begin{figure}[ht]
% \vspace{-2em}
    \centering
    \begin{subfigure}{0.35\linewidth}
    \centering
    \includegraphics[width=\linewidth]{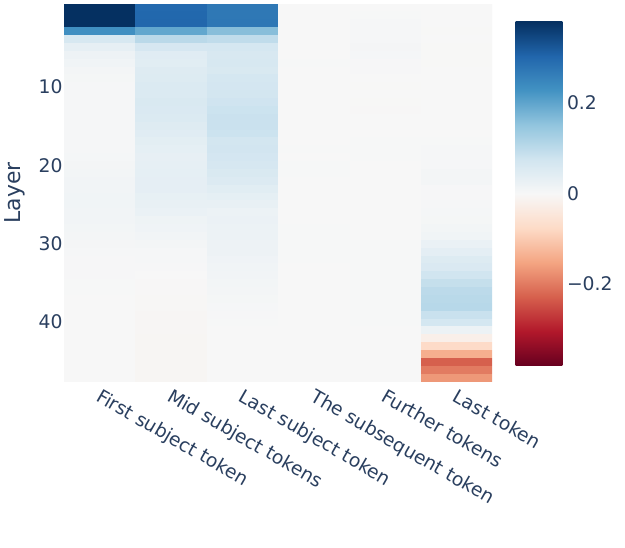}
    \caption{Logit difference (STR)}
    \end{subfigure}
    \qquad\qquad
    \begin{subfigure}{0.35\linewidth}
    \centering
        \includegraphics[width=\linewidth]{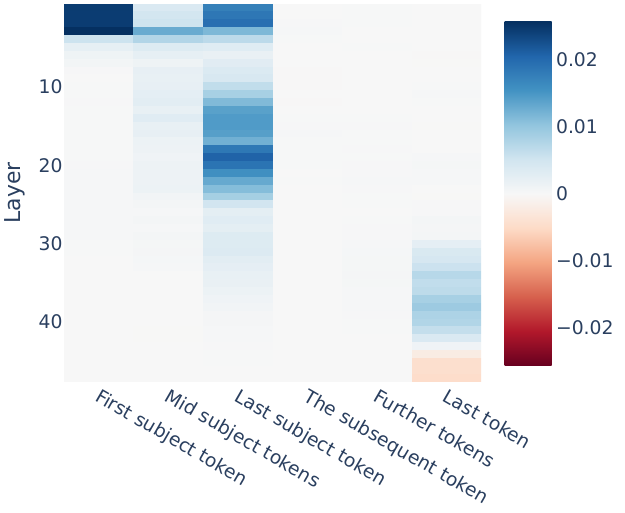}
        \caption{Probability (STR)}
    \end{subfigure}    
    \caption{\textbf{Activation patching on MLP} across layers and token positions in GPT-2 XL, with a sliding window patching of size $5$. Note that probability (b) highlights the importance of the last subject token, whereas logit difference (a) displays less effects.}
    \label{fig:rome-full}
\end{figure}

\paragraph{Experimental results}  For STR and window size of $5$, we plot  the patching effects across layers and positions in \autoref{fig:rome-full}. The visualization shows that probability assigns stronger effects at the last subject token than logit difference. Specifically, we calculate the ratio between   the sum of effects  (over all layers)  on the last subject token and those on the middle subject tokens. In both corruptions, probability assigns more effects to the last subject token than logit difference:
\begin{itemize}[leftmargin=1em]
    \item Using STR corruption, the ratio is $4.33$\texttimes\   in probability $>$ $1.22$\texttimes\ in logit difference.
    \item Using GN corruption, the ratio is $1.74$\texttimes\ in probability $>$ $0.77$\texttimes\ in logit difference. 
\end{itemize} 
This observation holds for other window sizes, too, for which we provide details in  Appendix \ref{sec:b2}.
We also validate our findings on GPT-J $6$B \citep{gpt-j} in Appendix \ref{sec:gpt-j}.
The results show that the choice of evaluation metrics influences the patching effects across tokens.
\subsection{Circuit discovery with probability}\label{sec:prob}
\cite{wang2022interpretability} discovers two Negative Name Mover (NNM) heads, 10.7 and 11.10, that noticeably  hurt model performance on IOI. 
In our previous   experiments on STR, both  are  detected, except when using probability as the metric where 11.10 is overlooked.
In fact,   the patching effect of 11.10  under STR in probability is well   within 2 SD from the mean    (mean $0.003$, SD $0.015$, and 11.10 receives $-0.022$). Looking closely, the reason is simple:
 \begin{itemize}[leftmargin=1em]
     \item 
In the corrupted run of STR, the average probability of  outputting the original  IO is $0.03$. Hence,  the patching effect in probability, $\P_{\text{pt}}(\text{IO})- \P_*(\text{IO})$,  is at least $ -0.03$, as $\P_{\text{pt}}(\text{IO})$ is non-negative. This  is already close to 2 SD below the mean ($-0.027$). Hence, for an NNM to be detected via patching, its $\P_{\text{pt}}(\text{IO})$ needs to be near $0$, which may be hard to reach.
\item By contrast, under GN corruption, the average probability of IO is $0.13$. Intuitively, this makes a lot more space for NNMs to demonstrate their effects. 
 \end{itemize}
In general, probability must fail to detect negative model components, if corruption reduces the correct token probability to near  zero. We  now give a cleaner experimental demonstration of this concern, using an original approach of \cite{wang2022interpretability}.
\paragraph{Experimental setup}
We revisit an alternative corruption method
proposed by \cite{wang2022interpretability}, 
where  S1, S2 and IO are   replaced by three   unrelated random names\footnote{This corrupted distribution is denoted by $p_{\text{ABC}}$ in the original paper of \cite{wang2022interpretability}
}; 
for example, ``John and Mary [...], John'' $\rightarrow$ ``Alice and Bob [...], Carol.''
 We use  probability of the original IO as the metric. Intuitively, this replacement method would achieve much stronger corruption effect, since it removes all the relevant information (S and IO) of the original IOI sentence. 

\paragraph{Experimental results}
First, we observe that the probability of outputting the   IO of the original IOI sentence  is negligible ($5\mathrm{e}{-4}$) under this corruption. As a result, using probability   detects neither NNMs. On the other hand, we find that logit difference still can. See Appendix \ref{sec:full-random} for the plots. In \autoref{sec:which-to},  we   confirm the same finding when corruption is applied to S1 and IO only.

At a high-level, we believe that this is a pitfall of  probability as an evaluation metric. Its non-negative nature  makes it incapable of discovering negative model components in certain settings.

\section{Sliding window patching}
\label{sec:sliding}
In this section, we examine the technique of sliding window patching in localizing factual information \citep{meng2022locating}.  For each layer, the method patches multiple adjacent layers simultaneously and computes the joint effects.
Hence, one should interpret the result of \cite{meng2022locating} as the effects being constrained {within a window}   rather than    at a single layer. 
We   argue that such as hypothesis can be tested by an alternative approach and we compare the results from these two.
\paragraph{Experimental setup} 
Instead of restoring multiple layers simultaneously,
we   patch  each individual MLP layer  one at a time. 
Then as an aggregation step, for each layer,   sum up the single-layer patching effects of its adjacent layers. For example, we add up the   effect at layer 2 to layer 6  to get an aggregated   effect for layer $4$. We patch the MLP  output at the last subject token. %and ary the  window sizes to be $3,5,10$. 
\paragraph{Experimental results}
\begin{figure}[t]
\vspace{-2em}
    \centering
    \begin{subfigure}[t]{0.23\linewidth}
    \includegraphics[width=\linewidth]{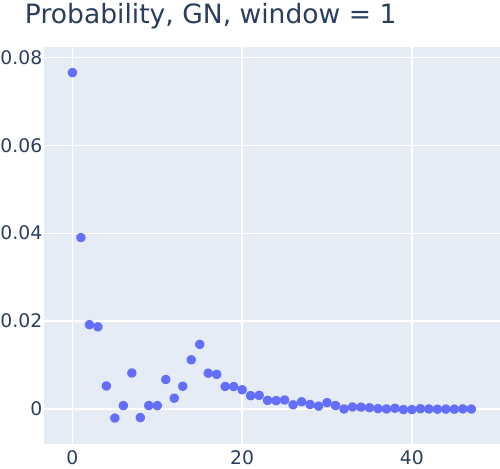}
    \caption{Single-layer patching}
    \label{fig:single-layer-main}
    \end{subfigure}
    \begin{subfigure}[t]{0.228\linewidth}
    \includegraphics[width=\linewidth]{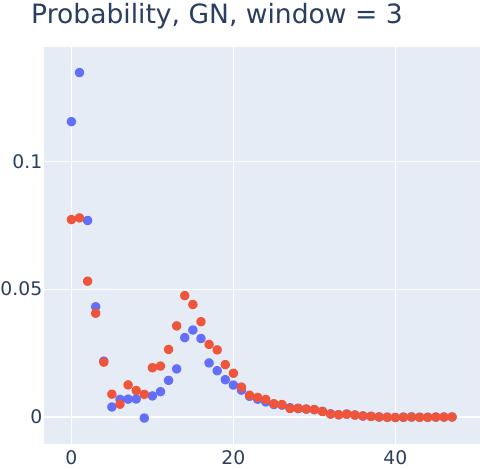}
    \caption{Window size of 3}
    \end{subfigure}
    \begin{subfigure}[t]{0.228\linewidth}
        \includegraphics[width=\linewidth]{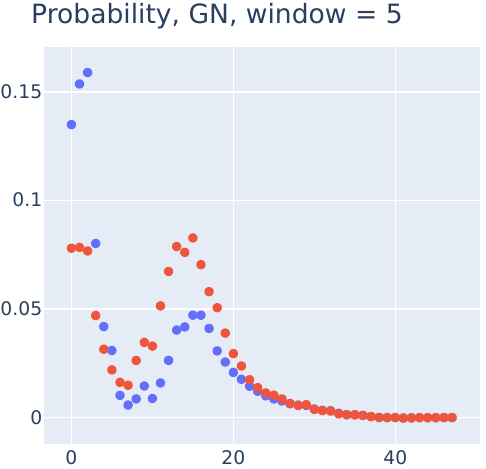}
        \caption{Window size of 5}
        \label{fig:sliding-5-main}
    \end{subfigure}
     \begin{subfigure}[t]{0.292\linewidth}
        \includegraphics[width=\linewidth]{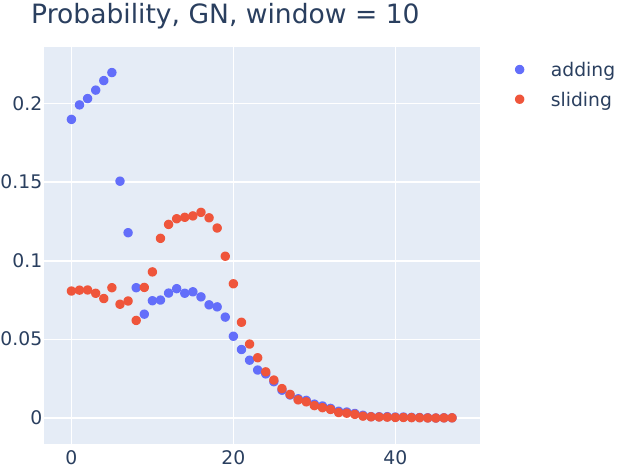}
        \caption{Window size of 10}
    \end{subfigure}
    \caption{\textbf{Sliding window patching vs summing up individual patching effects}; patching MLP activation at the last subject token in GPT-2 XL on factual recall prompts. Sliding window patching offers $1.40$\texttimes, $1.75$\texttimes\ and $1.59$\texttimes\ peak value than summation of single-layer patchings.   Single-layer patching (a) suggests a weak peak.}
    \label{fig:main-sliding}
\end{figure}
For each window size, we compute the ratio of the maximum  patching effect at the middle MLP layers between sliding window patching and summation of single-layer patching.  Over the combinations of window sizes, metrics and corruption methods, we find sliding window patching typically  provides   at least $20\%$ more peak effect than the summation method.

In \autoref{fig:main-sliding}, for window sizes of $3,5,10$,
we plot 
the results using GN corruption and probability as the metric, the original setting as in \cite{meng2022locating}. We observe significant gaps between the sliding window and the summation method. Moreover, for single-layer patching, the peak at layer 15 is fairly weak (\autoref{fig:single-layer-main}). Sliding window patching appears to generate more pronounced the concentration, as we increase the window sizes.

The result  suggests that sliding window patching tends to amplify   weak localization from single-layer patching (see \autoref{fig:single-rome} for plots on single-layer MLP patching in GPT-2 XL).  We believe this may arise due to  certain non-linear effects in joint patching and therefore results from which should be carefully interpreted;   see  \autoref{sec:discussion} for more discussions. 

\section{Discussion and recommendations}\label{sec:discussion}
We have observed  a variety of gaps between corruption methods and evaluation metrics used in activation patching on language models.
In this section, we summarize our findings and provide recommendations.

\paragraph{Corruption methods} 
We are concerned that GN corruption puts the model off distribution by introducing noise never seen during training. 
Indeed, in \autoref{sec:ood}, we provide evidence that in the corrupted run,  model's internal functioning is OOD relative to the clean distribution. 
This may induce unexpected anomalies in the model behavior, interfering with our ability to localize behavior to specific components. 
Conceivably,  GN corruption could  even lead to unreliable or    illusory results.
 
More broadly, this presents  a challenge to any intervention techniques that introduce OOD inputs to the model or its internal layers, including   ablations.
In fact, similar concerns have been raised earlier in the  interpretability   literature on feature attribution as well; see e.g. \cite{hooker2019benchmark,janzing2020feature, hase2021out}.

In contrast, STR provides counterfactual prompts (``The Eiffel Tower is in'' vs ``The Colosseum is in'') that are in-distribution and thus induces in-distribution activations,  avoiding the OOD issue. Therefore, we recommend STR whenever possible. GN may be considered as an alternative when token alignment or lack of analogous tokens makes STR unsuitable. 

\paragraph{Evaluation metrics} We generally recommend avoiding using probability as the metric, given that it may fail to detect negative model components. 

We find logit difference a convincing metric for localization in language models. Consider  an IOI setting where a model   contains an attention head that boosts the logits of all (single-token) names.  This head, though important, should   not be viewed as   part of the IOI circuit, but   our interventions may still   affect it.\footnote{We note that if our interventions do not affect the head, then it will not show up on any metric.} By measuring $\text{Logit}(\text{IO})-\text{Logit}(\text{S})$, logit difference controls for such   components and ensures they are not detected. This may not be achieved by other metrics, such as   probability or Logit(IO) alone.

KL divergence tracks the full model output distributions, rather than focused only on the correct or incorrect answer, and can be  a reasonable metric for circuit discovery as well \citep{conmy2023towards}.

\paragraph{Sliding window patching}
We speculate that simultaneously patching multiple layers could capture  the following non-linear effects and results in inflated localization plots: 
 \begin{itemize}[leftmargin=1em]
     \item 
Joint patching may suppress the flow of corrupted information within the  window of patched layers,  where single-layer patching offers no such control.
\item 
A window of patched layers may jointly perform a crucial piece of computation, such as  a major boost to the logit of the correct token, which no individual layer can single-handedly achieve.
 \end{itemize}
Generally, when examining the outcome from sliding window patching, one should be aware of the possibility of multiple layers working together.  Thus, the results  from the technique are to be interpreted as the joint effects of the full window, rather than of a single layer.
In practice, we recommend  experimenting with single-layer patching first and only consider sliding window patching when individual layers seem to induce small effects.  

\paragraph{Which tokens to corrupt?} 
In some problem settings, a prompt contains multiple key tokens, all relevant to completing the task. This would offer the flexibility to choose which tokens to corrupt. This is another important dimension of activation patching. For instance,   our   experiments  on IOI  in \autoref{sec:corruption} corrupt the S2 token. An alternative is to corrupt the S1 and IO. While this may seem an implementation detail, we find that this can greatly affect the localization outcomes. 

Specifically, in \autoref{sec:which-to}, we  test   corrupting S1 and IO in activation patching on IOI sentences,  by changing their values to random names or adding noise to the token embeddings . We find that almost all techniques   discover   the $3$ Name Mover (NM) Heads of the IOI circuit (\autoref{tab:ioi-io-s1} and \autoref{fig:ioi-xya-str}). These are attention heads that directly contribute to Logit(IO) as shown by \cite{wang2022interpretability}.
In contrast, our prior experiments corrupting S2  miss most of them (\autoref{tab:ioi}).

We intuit  that corrupting different tokens allows activation patching to trace different information within the model, thereby suggesting varying localizations results. 
For instance, in our prior experiments replacing S2 by IO, patching traces the value of IO or its position. On the other hand, in changing the values of S1 and IO while fixing their positions, patching highlights exactly where the model processes these values. 

In practice, we recommend trying out different     tokens to corrupt when the problem setting offers such flexibility. This may lead to   more exhaustive circuit discovery.

 \section{Related work}
 \paragraph{Activation patching}
 Activation patching is a variant of causal mediation analysis \citep{vig2020investigating,pearl2001direct}, similar forms of which are used broadly   in the interpretability literature \citep{soulos2020discovering,geiger2020neural,finlayson2021causal, geiger2022inducing}.  The specific one with GN corruption was 
first proposed by \cite{meng2022locating} under the name of causal tracing. \cite{wang2022interpretability, goldowsky2023localizing}  generalize this to a more sophisticated version of  path patching.  

\paragraph{Circuit analysis} Circuit analysis provides post-hoc model interpretability  \citep{casper2022sok}. This line of work is inspired by \cite{cammarata2020thread,elhage2021mathematical}.
Other works include \cite{geva2022transformer,li2022emergent,nanda2022progress,chughtai2023toy,zhong2023clock,nanda2023emergent,varma2023explaining,wen2023interpretability,hanna2023does,lieberum2023does}. Circuit analysis often requires  manual   effort by researchers, motivating recent work to scale or automate parts of the  workflow \citep{chan2022causal,bills2023language,conmy2023towards,geiger2023finding,wu2023interpretability,lepori2023neurosurgeon}. 

\paragraph{Mechanistic interpretability (MI)} MI aims to explain the internal computations and
representations of a model. 
While circuit analysis is a major direction under this broad theme, other recent case studies of MI in language model include \cite{mu2020compositional,geva2021transformer,yun-etal-2021-transformer,olsson2022context,scherlis2022polysemanticity,dai2022knowledge,gurnee2023finding,merullo2023language,mcgrath2023hydra,Bansal2023, dar2023analyzing, li2023transformers,brown2023privileged,katz2023interpreting, cunningham2023sparse}.

\section{Conclusion}\label{sec:conclusion}
We examine the role of metrics and methods in  activation patching in language models. We find that variations in these techniques could lead to different interpretability  results. We provide several recommendations towards the best practice, including the use of STR as the corruption method. 

% We see our work as a first step towards clarifying the 
% impact of hyperparameters in activation patching. We believe that this is important for understanding the robustness of the interpretability claims  drawn from applications of  this technique.

In terms of limitations,   our experiments are on decoder-only language models of   size up to $6$B. We leave it as a future direction to study other architectures and even larger models. Our work tests overriding corrupted activations by clean activations.    
The other direction---patching corrupted to clean---has also been used for circuit discovery, and  it is interesting to compare these two. 
In addition, we provide tentative evidence that certain corruption methods   lead to OOD model behaviors and suspect that this can make the resulting interpretability claims  unreliable.
Future work should examine this hypothesis closely and furnish further  demonstrations. 
Finally, it is interesting to develop   more principled techniques   for activation patching or  propose other    methods for localization.
 \subsubsection*{Acknowledgments}
FZ would like to thank  Matthew Farhbach, Dan Friedman, Johannes Gasteiger, Asma Ghandeharioun, Stefan Heimersheim,  János Kramár, Kaifeng Lyu, Vahab Mirrokni,   Jacob Steinhardt and Peilin Zhong for helpful discussions, and Jiahai Feng, Yossi Gandelsman, Oscar Li and Alex Wei for comments on  early drafts of the paper.

\bibliography{iclr2024_conference}
\bibliographystyle{iclr2024_conference}

\appendix
\section{Review of Transformer Architecture}
\label{sec:arc}
We follow the notation of \cite{elhage2021mathematical} and give a review of the Transformer architecture \citep{vaswani2017attention}. The input $x_0\in \R^{N\times d}$  to a  transformer model is a sum of position and token embeddings, where $N$ is the sequence length and $d$ is the dimension of the model's internal states.  The input is the initial value of the residual stream which subsequently gets updated by the transformer blocks. 

Each transformer block consists of a  multi-head self-attention sublayer and an  MLP sublayer. (For GPT-J, these two sublayers are   parallelized.) The MLP sublayer is a two-layer feedforward network that processes each token position independently in parallel. Following \cite{elhage2021mathematical}, the output of the attention sublayer can be decomposed into individual heads. For the $i$th layer, the attention output can be written as  $y_i=\sum_{j=1}^H h_{i, j}\left(x_i\right)$, where $h_{i,j}$ denotes the $j$th attention head of the layer. Each head has four weight matrices, $W^{i,j}_Q, W^{i,j}_K,W^{i,j}_V\in \mathbb{R}^{d \times \frac{d}{H}}$ and $W_O \in \mathbb{R}^{\frac{d}{H}\times d}$.  For a residual stream $x$, we refer to $Q^{i,j}=x W^{i,j}_Q, K^{i,j} = x W^{i,j}_K, V^{i,j}= x W^{i,j}_V$ as the query, key and value of the head. The attention pattern is given by 
\begin{align*}
    A^{i,j} = \text{softmax} \left (  \frac{(x W^{i,j}_Q)( x W^{i,j}_K)^T}{\sqrt{d/H}} + M  \right) \in \R^{N\times N},
\end{align*}
where $M$ is the attention mask. In auto-regressive language models, the attention pattern is  masked to a lower triangular matrix.  The output of the attention sublayer is given  by 
\begin{equation}
x + \text{Concat}\left[A^{i,1} V^{i,1}, \ldots, A^{i,j} V^{i,j}, \ldots, A^{i,H} V^{i,H}\right] W_O.
\end{equation}
\section{Details on Experimental Settings}
\label{sec:exp-details}
For Gaussian noise (GN) corruption, we corrupt the embeddings of the crucial tokens by adding a Gaussian
noise $\varepsilon \sim \mathcal{N}(0 ; \nu)$,  where $\nu$ is set to be $3$ times the   standard deviation of the token embeddings from the dataset \citep{meng2022locating}.

We always perform GN and STR experiments in parallel. For STR, there is a natural the incorrect token $r'$, since $\xco$ is also a valid in-distribution prompt. This  allows for a well-defined metric of logit difference $\LD(r,r')=\text{Logit}(r) - \text{Logit}(r')$.
To make a fair comparison, the same $r'$ is used for evaluating the logit difference metric under GN. 

Throughout the paper, layers are zero-indexed, numbered from $0$ to $L - 1$ rather than $1$ to $L$. 
\paragraph{Factual recall}
To perform STR in the factual association setting, we construct \textsc{PairedFacts}, a dataset of $145$ pairs of  prompts. Within each pair, the two prompts have the same sequence length (under the GPT-2 tokenizer) but distinct answers. All the prompts are selected from the \textsc{CounterFact} and \textsc{Known1000} datasets of \cite{meng2022locating}. On these prompts, 
\begin{itemize}[leftmargin=1em]
    \item 
GPT-2 XL achieves an average of $49.0\%$ probability on the correct token and $6.85$ logit difference.
\item GPT-2 large achieves $41.1\%$ and $5.88$ logit difference.
\item GPT-J achieves $50.1\%$ and $7.36$ logit difference.
\end{itemize}
 A few samples of the \textsc{PairedFacts} dataset are listed in \autoref{fig:sample-pairs} of \autoref{sec:dataset}. 

Since the prompts are perfectly symmetric and all of them are in-distribution, our  STR experiments consist of both ways, where a prompt within a pair play the  role  of both $\xco$ and $\xcl$. 

Our experiments with GN corruption is performed in the same manner as in \cite{meng2022locating, hase2023does}, with noise applied to all subject tokens' embeddings.

The experiments here are implemented via the TransformerLens library \citep{nandatransformerlens2022}.

\paragraph{IOI}
Unless specified otherwise, GN applies Gaussian noise to the S2 token embedding. Over $500$ prompts, the   probability of outputting IO is $\P_*(r) =0.129$ under GN corruption (with $r$ being IO), whereas it is $0.481$ under the clean distribution $p_{\text{IOI}}$.

All our experiments are performed using the original codebase of
\cite{wang2022interpretability},
available at {\small \url{https://github.com/redwoodresearch/Easy-Transformer}}.
The code provides the functionality of constructing $\xco$ under various definitions of corruptions, including STR.

\section{Results on arithmetic reasoning in GPT-J}
\label{sec:arithmetic}
\paragraph{Experimental setup}
We follow the setting of \cite{stolfo2023understanding} and perform localization analysis on the task of basic arithmetic in GPT-J \citep{gpt-j}, a  decoder-only model with $6$B parameters. For simplicity, we consider addition, subtraction and multiplication up to $3$ digits.  We provide the model with a $2$-shot prompts of the format 
\begin{align*}
    X_1&+Y_1=Z_1\\X_2&+Y_2=Z_3\\X_3&+Y_3=
\end{align*}
where the numbers $X_i,Y_i$ are random integers and the operator can be $+,-,\times$. \cite{stolfo2023understanding} finds that this leads to improved accuracy. Since large integers get split into multiple tokens, we draw $X_i,Y_i$ from $\{1,2,\cdots,250\}$ for addition and subtraction and from $\{1,2,\cdots,23\}$ for multiplication. 
To obtain a dataset for activation patching, we first draw $200$ prompts and discard those on which the model's top-ranked output token is incorrect. 

We set GN corruption to add noise to the token embeddings at the positions of $X_3, Y_3$. Similarly, STR replaces $X_3,Y_3$ by two random integers drawn from the same set, which ensures that the corrupted prompt is still in-distribution. We remark that  \cite{stolfo2023understanding} applies the same STR corruption in their patching experiments. 

\cite{stolfo2023understanding} devises a new metric to evaluate the patching effects. More precisely, they report:
\begin{align}\label{eqn:smetric}
    \frac{1}{2} \left[ \frac{\P_\text{pt} (r) - \P_*(r)}{\P_*(r)} +  \frac{\P_*(r') -\P_\text{pt} (r') }{\P_\text{pt} (r')} \right]
\end{align}
from patching the MLP activation at last token of the prompt.\footnote{Note that our notations here are different from \cite{stolfo2023understanding}.} We compute the patching effect given by the metric, as well as probability and logit difference.

\begin{figure}[ht]
    \centering
    \begin{subfigure}{0.32\linewidth}
    \centering
    \includegraphics[width=\linewidth]{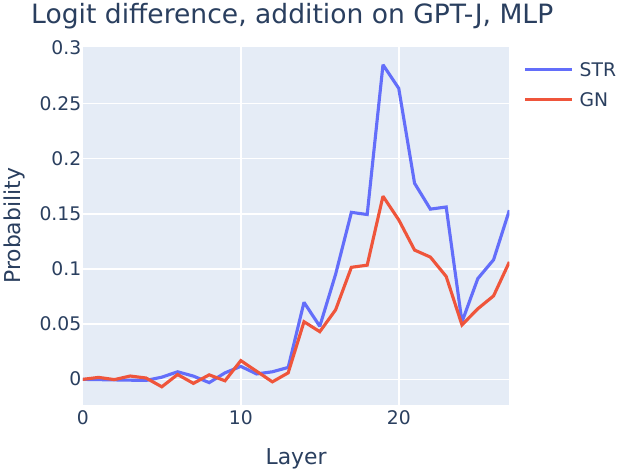}
    \caption{Logit difference as the metric}
    \end{subfigure}
    \begin{subfigure}{0.32\linewidth}
    \centering
        \includegraphics[width=\linewidth]{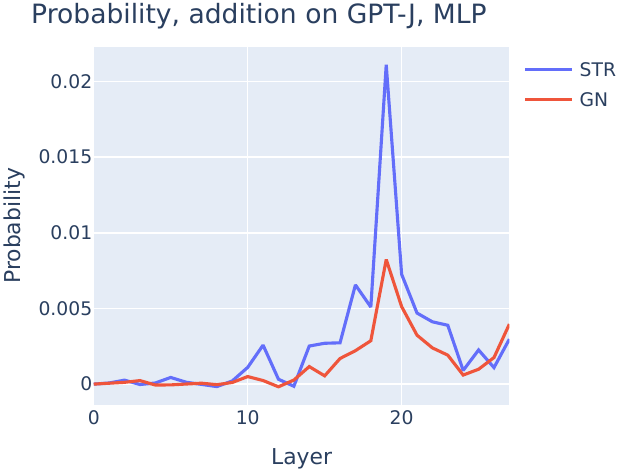}
        \caption{Probability as the metric}
    \end{subfigure}
    \begin{subfigure}{0.32\linewidth}
    \centering
        \includegraphics[width=\linewidth]{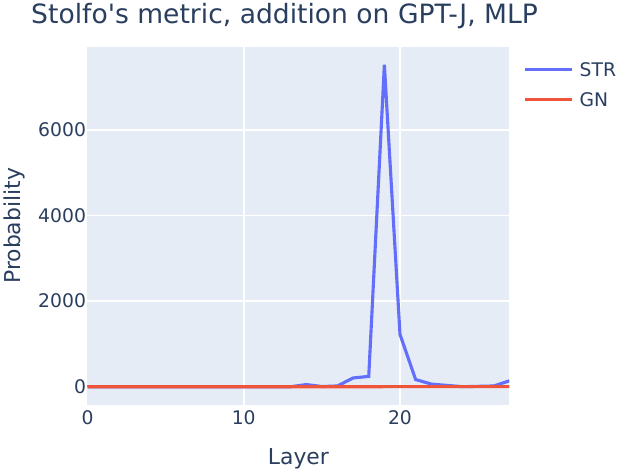}
        \caption{\cite{stolfo2023understanding}'s metric}
    \end{subfigure}
    \caption{\textbf{The effects of patching MLP layers} in GPT-J on addition prompts.}
    \label{fig:add-j}
\end{figure}

\begin{figure}[ht]
    \centering
    \begin{subfigure}{0.32\linewidth}
    \centering
    \includegraphics[width=\linewidth]{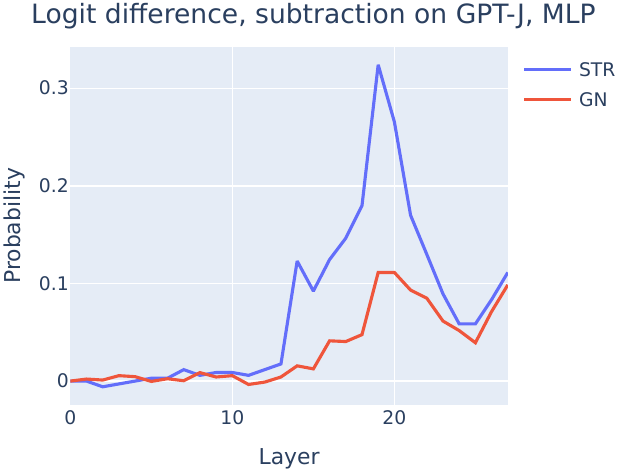}
    \caption{Logit difference as the metric}
    \end{subfigure}
    \begin{subfigure}{0.32\linewidth}
    \centering
        \includegraphics[width=\linewidth]{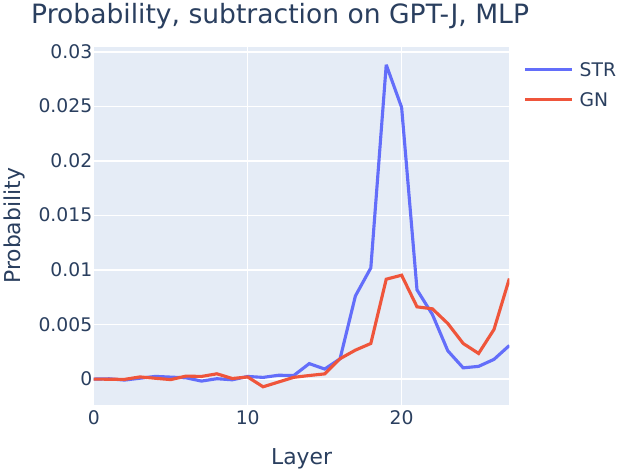}
        \caption{Probability as the metric}
    \end{subfigure}
    \begin{subfigure}{0.32\linewidth}
    \centering
        \includegraphics[width=\linewidth]{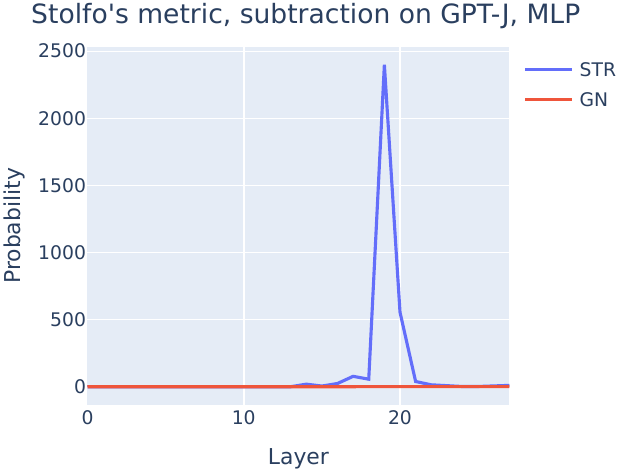}
        \caption{\cite{stolfo2023understanding}'s metric}
    \end{subfigure}
    \caption{\textbf{The effects of patching MLP layers} in GPT-J on subtraction prompts.}
    \label{fig:sub-j}
\end{figure}

\begin{figure}[ht]
    \centering
    \begin{subfigure}{0.32\linewidth}
    \centering
    \includegraphics[width=\linewidth]{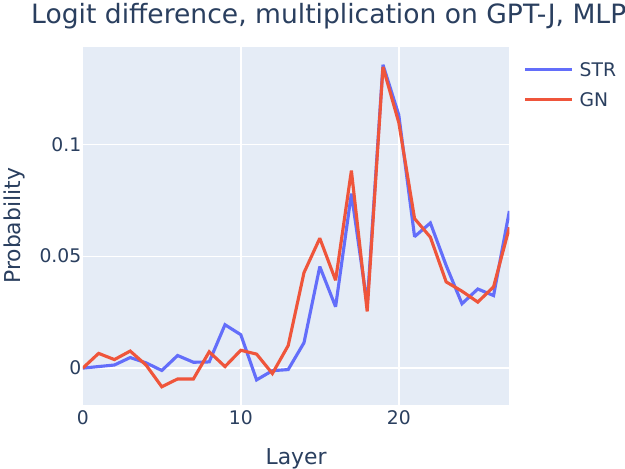}
    \caption{Logit difference as the metric}
    \end{subfigure}
    \begin{subfigure}{0.32\linewidth}
    \centering
        \includegraphics[width=\linewidth]{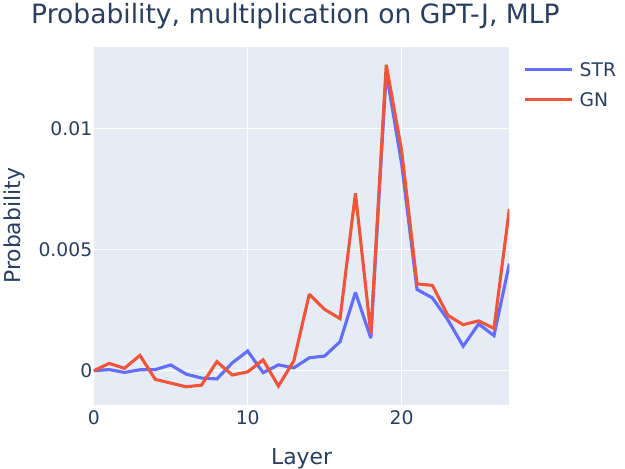}
        \caption{Probability as the metric}
    \end{subfigure}
    \begin{subfigure}{0.32\linewidth}
    \centering
        \includegraphics[width=\linewidth]{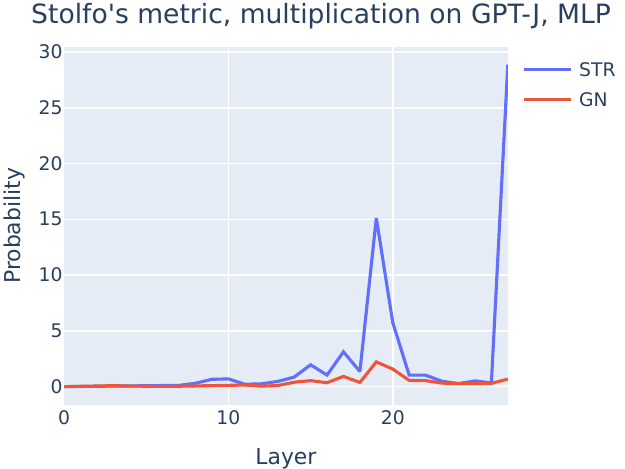}
        \caption{\cite{stolfo2023understanding}'s metric}
    \end{subfigure}
    \caption{\textbf{The effects of patching MLP layers} in GPT-J on multiplication prompts.}
    \label{fig:mult-j}
\end{figure}

Following \cite{stolfo2023understanding}, we narrow our focus on localization of MLP layers.
All the experiments  patch a single MLP layer's activation at the last token of the prompt.
\paragraph{Experimental results} 
Focused on the logit difference and probability metric, we   observe gaps between GN and STR for addition and subtraction.  In particular, STR is found to provide sharper concentration, up to a magnitude of $4$\texttimes. This in contrast with our results on factual association (\autoref{sec:corruption-diff}), where GN appears to induce stronger peak. For multiplication, GN and STR provides nearly matching results.  This highlights that     activation patching can be sensitive to corruption methods in a rather unpredictable way. See \autoref{fig:add-j}, \autoref{fig:sub-j} and \autoref{fig:mult-j} for plots.

For the metric (\ref{eqn:smetric}) of \cite{stolfo2023understanding}, we qualitatively replicate their results, similar to Figure 2 of their paper, and  find extremely pronounced peak with STR corruption. Towards understanding this observation,  we examine the quantity (\ref{eqn:smetric}) closely and discover that its first term typically dominates the second. This, in turn, is because the denominator term $\P_*(r)$, the probability of outputting the correct answer in the corrupted run, is usually tiny under STR corruption. 
The small denominator, therefore, acts as a large multiplier that amplifies the absolute gap between patching different layers. We note that this effect is much smaller under GN since  $\P_*(r)$ is usually not negligible.


\section{Results on Python docstring circuit} 
\label{sec:docstring}
\cite{docstring} 
studies a circuit for 
Python docstring completion in a pre-trained $4$-layer attention-only Transformer.\footnote{The model is available in the TransformerLens library under the name \texttt{attn-only-4l} \citep{nandatransformerlens2022}.}  We do not aim to fully replicate their results. Rather, we perform  patching experiments to localize the important attention heads for the purpose of evaluating variants of  activation patching.

\paragraph{Experimental setup}
In \cite{docstring}, a Python docstring completion instance consists of the following prompt:
\begin{lstlisting}[
    basicstyle=\ttfamily\small
]
def rand0(self, rand1, rand2, A_def, B_def, C_def, rand3):
    """rand4 rand5 rand6

    :param A_def: rand7 rand8
    :param B_def: rand9 rand10
    :param
\end{lstlisting}
where \texttt{rand}'s are random single-token English words and the goal is to complete the prompt with \texttt{C\_def}.
\cite{docstring} finds that the $4$-layer model solves the docstring task with an   accuracy of 56\% and the   logit difference is $0.5$. 

Following their approach, we run activation patching on all attention heads, across all token positions. This is more fine-grained than what we did for the IOI circuit, since the outcome would also highlight the token positions that matter for the important heads. 

We apply corruption to the  \texttt{C\_def} token. For STR, it is replaced randomly by a single-token English word in the same way specified in \cite{docstring}.  
\paragraph{Experimental results}
We take $200$ instances of  the docstring completion task, perform activation patching  by positions and compute the patching effects. We report all position-head pairs with patching effect   $2$ standard deviations away  from the mean. We find that the detections are mostly at the position of \texttt{C\_def} and the last token of the prompt. The details are given in \autoref{tab:docstring}.
\begin{table}[ht]
\centering
\begin{tabular}{l|l||c|c}
\toprule
\textbf{Corruption}  & \textbf{Metric}      & At the position of \texttt{C\_def}&  \multicolumn{1}{c}{At the last position}   \\ \midrule
STR               & Logit difference          &      0.0, 0.1,   0.5    &   2.3, 3.0, 3.5, 3.6           \\ 
STR               & Probability &     &3.0, 3.6                      \\  
STR               & KL divergence          &  0.0, 0.1,   0.5, 2.2    &          2.3, 3.0, 3.6  \\ \midrule
GN $\dagger$                &Logit difference          &  0.5      &    1.4, 1.5, 2.2, 2.3, 3.0,   3.6, 3.7             \\  
GN              & Probability &      &       3.0, 3.6 \\ 
GN               & KL divergence &     &2.2, 2.3, 3.0, 3.5, 3.6     \\ 
\bottomrule
\end{tabular}
\caption{\textbf{Detections from activation patching} of attention heads by position on the Python docstring completion task. $\dagger$ {\small Also detects two early-layer heads active at other positions and  four negative heads active at the last position, which we omit here.}}
\label{tab:docstring}
\end{table}

We again find that the localization outcomes are sensitive to the choice of corruption method and evaluation metric. The results of GN appear quite noisy, except when using probability as the metric. On the other hand, we remark that 3.0 and 3.6 are consistently highlighted across metrics and methods. In fact, they are typically assigned  the largest patching effects (at the last position). This appears consistent with the result of \cite{docstring}, where 3.0 and 3.6 are found to be directly responsible for moving the \texttt{C\_def} token.
\section{Results on the greater-than circuit in GPT-2 small}
\label{sec:gt}
\cite{hanna2023does}
In this section, we study the greater-tan task, specified below, and perform activation patching on the attention heads in GPT-2 small. In this setting, the prior work by \cite{hanna2023does, conmy2023towards} show that model computation is fairly localized in this setting and provide a set of circuit discovery results. We remark that we do not attempt  to replicate the circuit discovery results here, but rather to evaluate whether activation helps with localizing certain important model components.

\paragraph{Experimental setup}
Following \cite{hanna2023does}, an instance of the greater-than task consists of an incomplete sentence of  the template: ``The \texttt{<noun>} lasted from the
year XXYY to the year XX'', where \texttt{<noun>} is a single-token word and XX and YY are two-digit numbers. For example, ``The war lasted from year 1745 to 17''. The goal is to complete the prompt with an integer greater than XX (in this case, 45). Across several metrics, \cite{hanna2023does} shows that GPT-2 small performs well on this task.  

We focus on the role of attention heads in our study. To perform corruption, we ensure that the year XXYY are tokenized as [XX][YY] by filtering out years and numbers that do not conform to the constraint. GN corruption adds noise to the token embedding of YY. Following \cite{hanna2023does,conmy2023towards}, STR corruption replaces YY by 01. The probability metric, in this setting, is defined as the sum of probabilities of the years greater than YY. The logit difference metric is defined as the sum of logits of the years greater than YY minus the sum of logits of the years less than YY. 

We perform activation patching on the attention heads outputs over all token positions. 

\paragraph{Experimental results}
We find significant difference between the results achieved by GN and STR. In fact, the set of heads that are localized by the methods are mostly disjoint. Specifically, GN appears to give extremely noisy results that are not in line with the findings of \cite{hanna2023does,conmy2023towards}. The details are given in \autoref{tab:gt}

\begin{table}[ht]
\centering
\begin{tabular}{l|l||c|c}
\toprule
\textbf{Corruption}  & \textbf{Metric}      &  Positive   &  \multicolumn{1}{c}{Negative}   \\ \midrule
STR               & Logit difference          &      6.9, 7.10, 8.11, 9.1,  10.4   &       \\ 
STR               & Probability &    7.10, 8.11, 9.1&            \\  
STR               & KL divergence          &     6.9, 7.10, 8.11, 9.1  &            10.7   \\ \midrule
GN   &Logit difference          &   0.9, 7.10, 8.10, 9.1, 10.4     & 6.1, 8.6, 9.5 \\  
GN              & Probability &  5.5, 6.1, 6.9, 7.10, 7.11, 8.8, 9.1   & \\ 
GN               & KL divergence & 5.5, 6.1, 6.9, 7.10, 7.11, 8.8, 9.1     &    5.9, 7.6    \\ 
\bottomrule
\end{tabular}
\caption{\textbf{Detections from activation patching  on attention heads for the greater-than task} in GPT-2 small, averaged across $300$ prompts.}
\label{tab:gt}
\end{table}

\begin{figure}[ht]
    \centering
    \begin{subfigure}{0.32\linewidth}
    \centering
    \includegraphics[width=\linewidth]{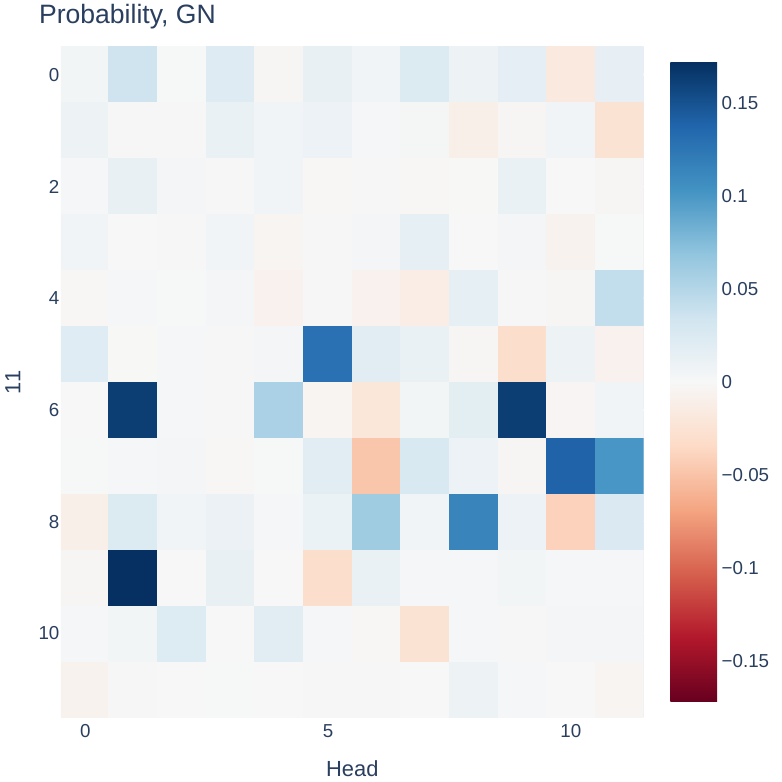}
    \caption{Probability as the metric}
    \end{subfigure}
    \begin{subfigure}{0.32\linewidth}
    \centering
        \includegraphics[width=\linewidth]{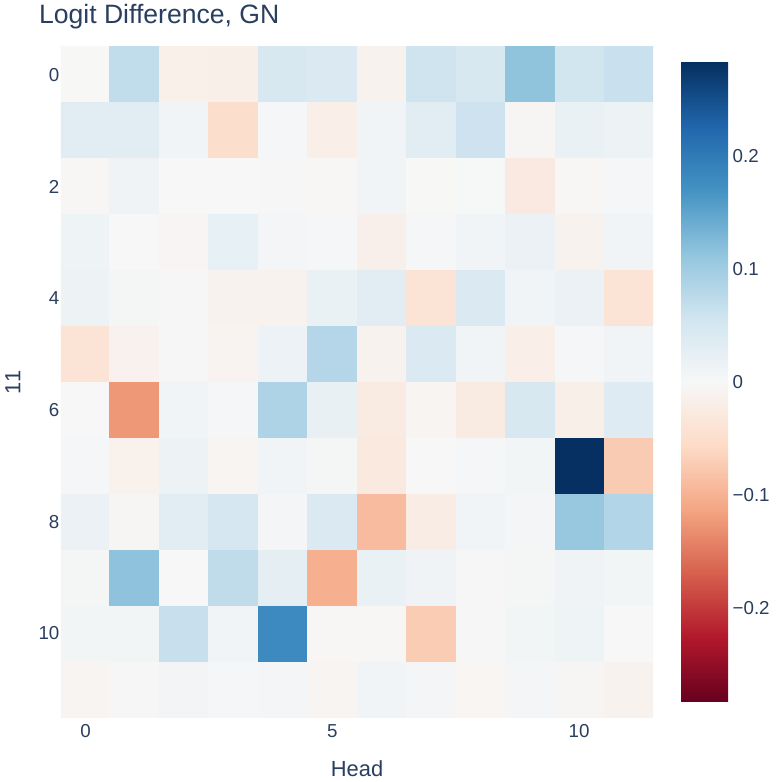}
        \caption{Logit difference as the metric}
    \end{subfigure}
     \begin{subfigure}{0.32\linewidth}
    \centering
        \includegraphics[width=\linewidth]{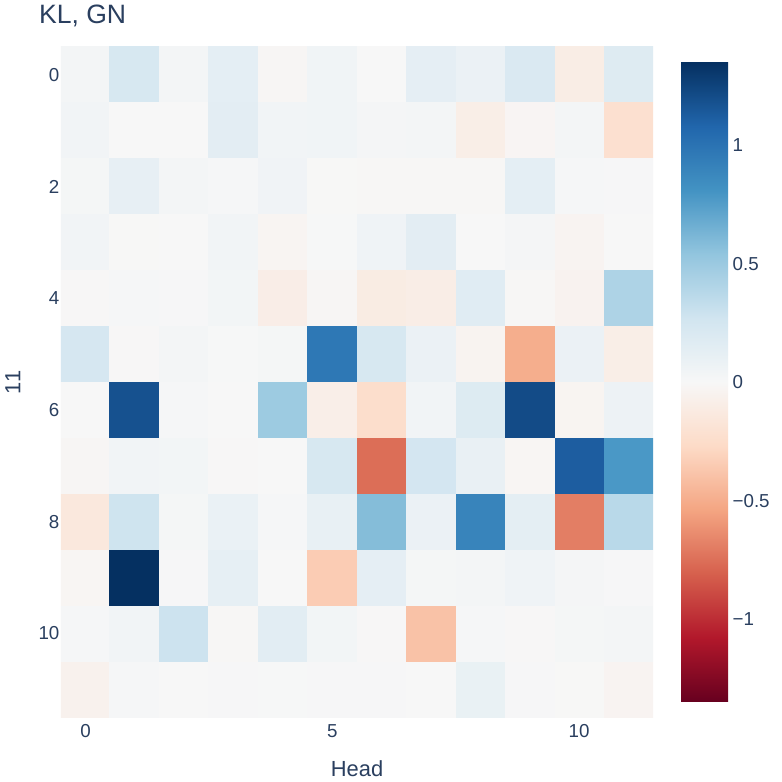}
        \caption{KL divergence as the metric}
    \end{subfigure}
    \caption{\textbf{The effects of patching attention heads} in GPT-2 small on the greater-than task, using GN corruption. We see that the results are fairly noisy and do not appear to be localized.}
    \label{fig:gn-gt}
\end{figure}

\begin{figure}[ht]
    \centering
    \begin{subfigure}{0.32\linewidth}
    \centering
    \includegraphics[width=\linewidth]{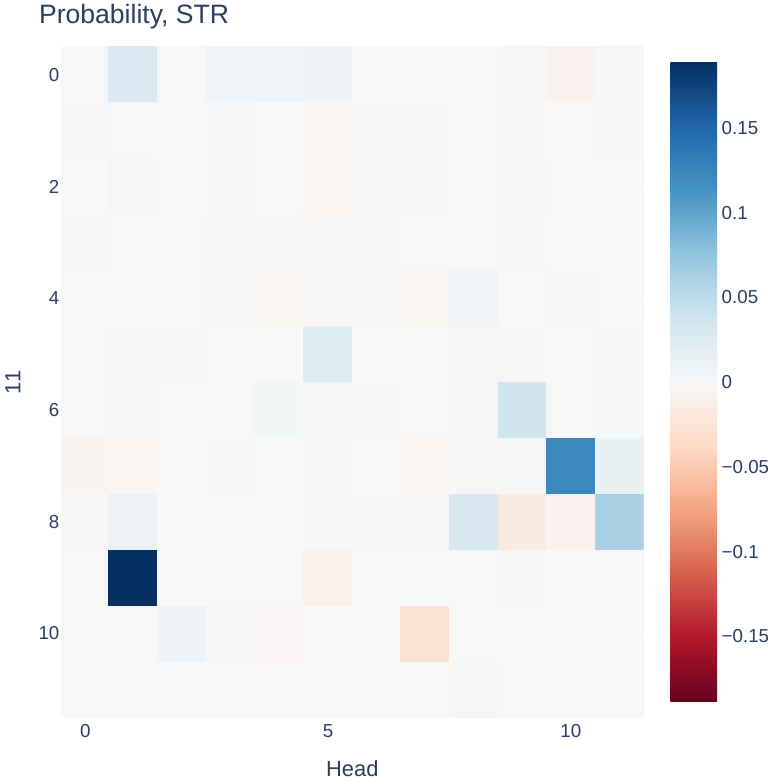}
    \caption{Probability as the metric}
    \end{subfigure}
    \begin{subfigure}{0.32\linewidth}
    \centering
        \includegraphics[width=\linewidth]{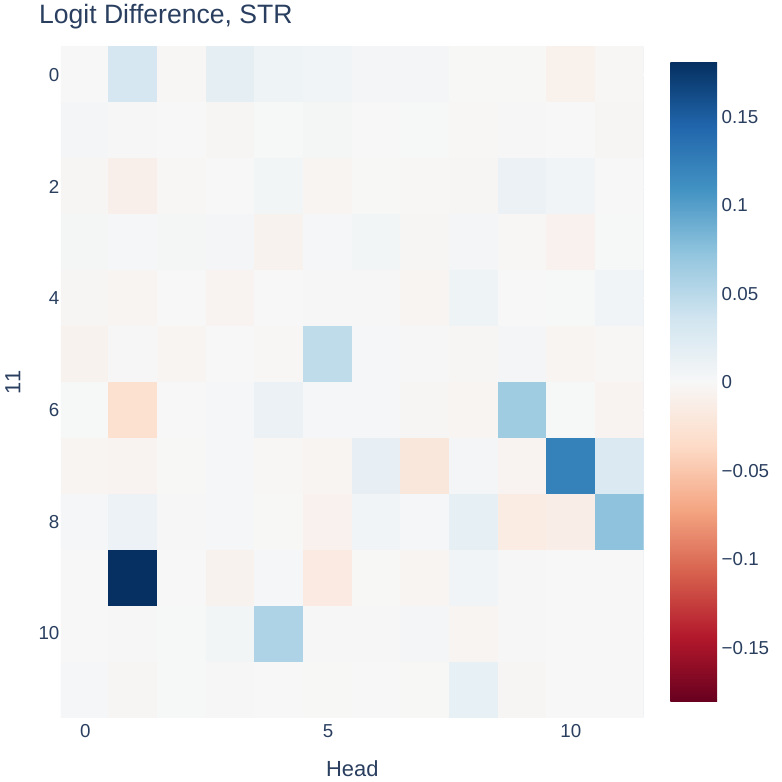}
        \caption{Logit difference as the metric}
    \end{subfigure}
     \begin{subfigure}{0.32\linewidth}
    \centering
        \includegraphics[width=\linewidth]{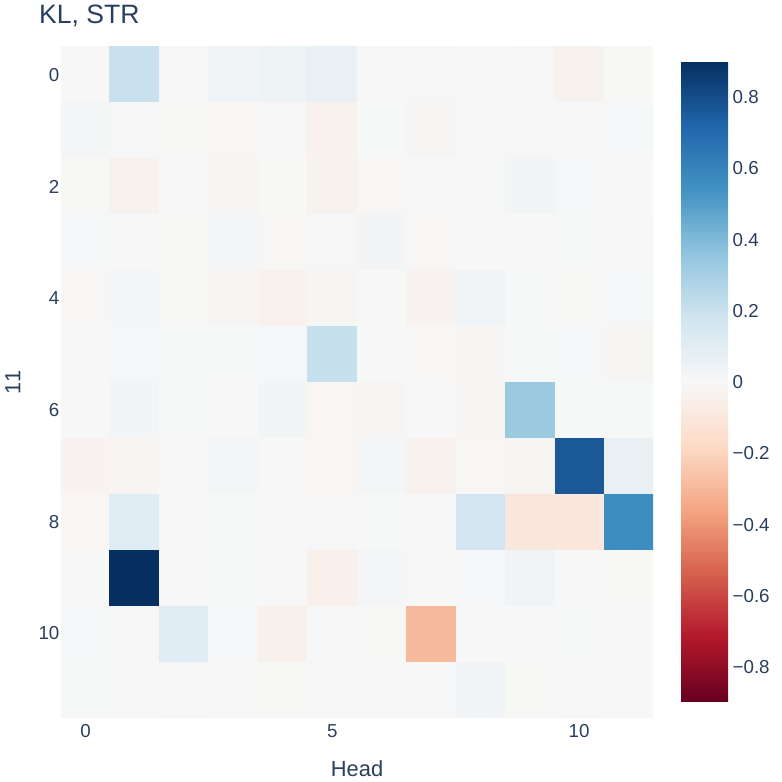}
        \caption{KL divergence as the metric}
    \end{subfigure}
    \caption{\textbf{The effects of patching attention heads} in GPT-2 small on the greater-than task, using STR corruption. This gives clearly localized results.}
    \label{fig:str-gt}
\end{figure}

The results from STR are fairly reasonable as the heads   6.9, 7.10, 8.11, 9.1  are also  discovered by \cite{hanna2023does, conmy2023towards}, using more sophisticated methods. In contrast, the heads discovered by GN corruption share little overlap with STR, except 7.10 and 9.1. From visualizations, we also see that the plots for GN experiments are fairly noisy and do not yield much localization at all (\autoref{fig:gn-gt}).  On the other hand, the plots from STR are easily interpretable (\autoref{fig:str-gt}).

\section{Which tokens to corrupt matters}
\label{sec:which-to}
In this section, we revisit the implementation of    corruption methods in the setting of IOI \citep{wang2022interpretability}.

Previously in our  STR  experiments in \autoref{sec:corruption} and \autoref{sec:metrics}, the S2 token was corrupted  by exchanging with IO. Similarly, in GN, we add noise to the token embedding of S2.
We notice that the localization results from this approach   miss at least $2$ out of the $3$ Name Mover (NM) Heads (\autoref{tab:ioi}); they directly contribute to the logit of IO as found by \cite{wang2022interpretability}. In particular, all combinations of metric and method would miss out on 9.6 and 10.0 (\autoref{tab:ioi-detection-list}).

We show that by varying exactly which tokens to corrupt, the NMs can be discovered, too

\paragraph{Experimental setup}
We consider the IOI setting \citep{wang2022interpretability} using STR and GN corruption for localizing attention heads. Here, we corrupt the S1 and IO tokens. For STR, we simply replace S1 and IO by two random unrelated names. In both STR and GN experiments, the S2 token remains the same as in $\xcl$.

We perform activation patching across all attention heads. We apply logit difference, probability and KL divergence as the metric. All the results are averaged across $500$ sampled IOI sentences. 

\paragraph{Experimental results} We find that most combinations of metrics and methods are able to notice all the NMs, when corruption applies to S1 and IO. We give the exact detections below and  categorize them into positive and negative for simplicity. 

\begin{table}[ht]
\centering
\begin{tabular}{l|l||c|c}
\toprule
\textbf{Corruption}  & \textbf{Metric}      &  Positive   &  \multicolumn{1}{c}{Negative}   \\ \midrule
STR               & Logit difference          &      9.6, 9.9, 10.0    &   10.7, 11.10     \\ 
STR               & Probability &    9.9 &            \\  
STR               & KL divergence          &   9.6, 9.9, 10.0    &           10.7, 11.10   \\ \midrule
GN   &Logit difference          &   9.6, 9.9, 10.0      &    10.7, 11.10          \\  
GN              & Probability &      9.6, 9.9, 10.0   & \\ 
GN               & KL divergence & 9.6, 9.9, 10.0    &   10.7, 11.10       \\ 
\bottomrule
\end{tabular}
\caption{\textbf{Detections from activation patching by corrupting S1 and IO}  in IOI. The Name Mover Heads are   9.6, 9.9, 10.0 and the Negative Name Mover Heads are 10.7 and 11.10, based on \cite{wang2022interpretability}. No other heads, including the S-Inhibition Heads, are noticed with this approach.}
\label{tab:ioi-io-s1}
\end{table}
First, we observe that this corruption seems to precisely target the  NMs and their Negative counterparts. Intuitively, this is natural. \cite{wang2022interpretability} finds that NMs write in the direction of the logit of the name (IO or S), whereas the Negative NMs do the opposite. 
Patching the clean activations of NMs recover such behavior. 

\begin{figure}[t]
    \centering
    \begin{subfigure}{0.32\linewidth}
    \centering
    \includegraphics[width=\linewidth]{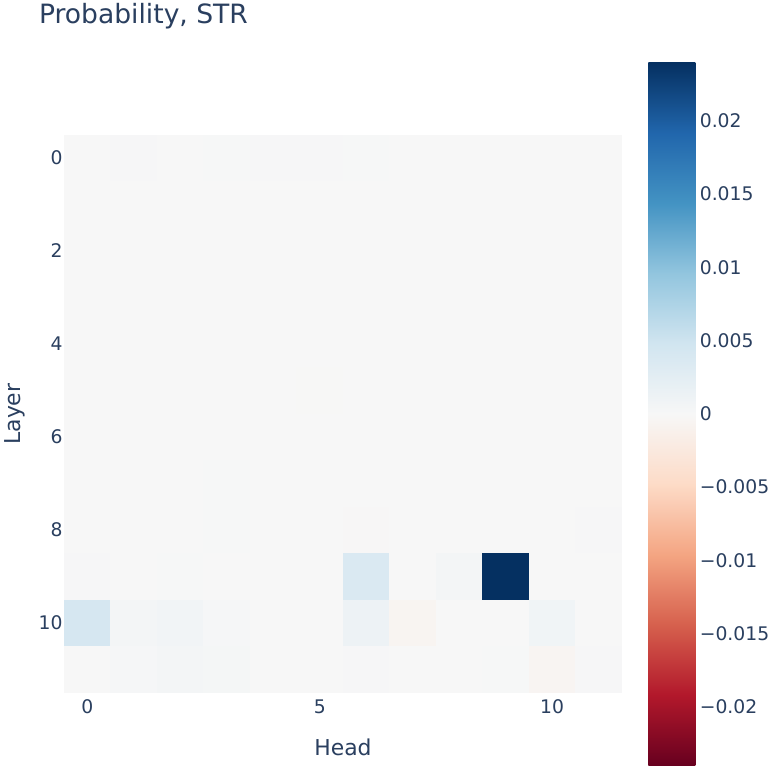}
    \caption{Probability as the metric}
    \end{subfigure}
    \begin{subfigure}{0.32\linewidth}
    \centering
        \includegraphics[width=\linewidth]{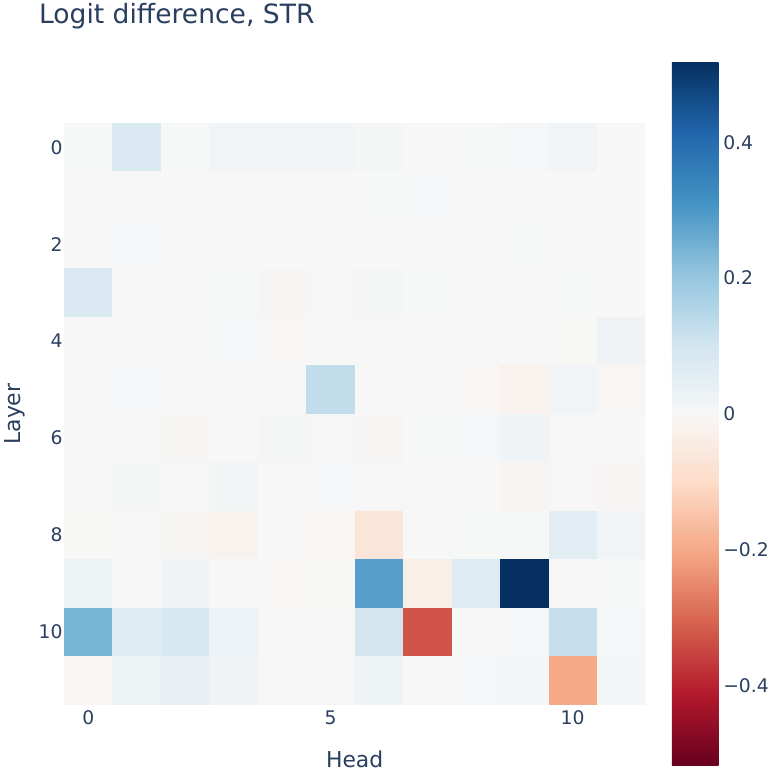}
        \caption{Logit difference as the metric}
    \end{subfigure}
     \begin{subfigure}{0.32\linewidth}
    \centering
        \includegraphics[width=\linewidth]{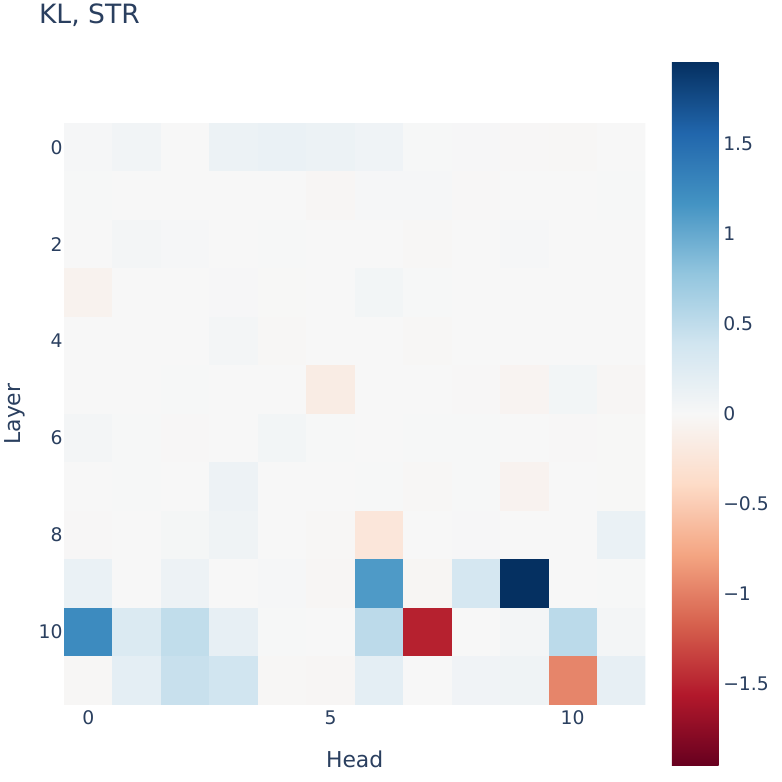}
        \caption{KL divergence as the metric}
    \end{subfigure}
    \caption{\textbf{The effects of patching attention heads} in GPT-2 small on IOI sentences, using STR corruption on S1 and IO.}
    \label{fig:ioi-xya-str}
\end{figure}

Second, we confirm our finding that probability will miss out on the Negative NM; see \autoref{fig:ioi-xya-str} for the plots.

Overall, the experiment suggests that exactly which token is corrupted affects the localization outcomes. Intuitively, varying the corrupted token(s) allows activation patching to trace different information within the model's computation paths; see \autoref{sec:discussion} for a discussion.

\section{Further details on  factual association}
\label{sec:further-fact}
The  plots of  subsection Appendix \ref{sec:b1} to \ref{sec:sliding-plots} are produced on GPT-2 XL and with the \textsc{PariedFacts} as dataset. Following that, we also experiment with the GPT-2 large \citep{radford2019language} and GPT-J \citep{gpt-j} model in Appendix \ref{sec:large-apx} and  \ref{sec:gpt-j}.

\subsection{Plots on MLP patching at the last subject token in GPT-2 XL}
\label{sec:b1}
\begin{figure}[ht]
    \centering
    \begin{subfigure}[t]{0.23\linewidth}
    \includegraphics[width=\linewidth]{img/rome/sliding-pr-gn-1.pdf}
    \caption{Probability (GN)}
    \end{subfigure}
    \begin{subfigure}[t]{0.23\linewidth}
    \includegraphics[width=\linewidth]{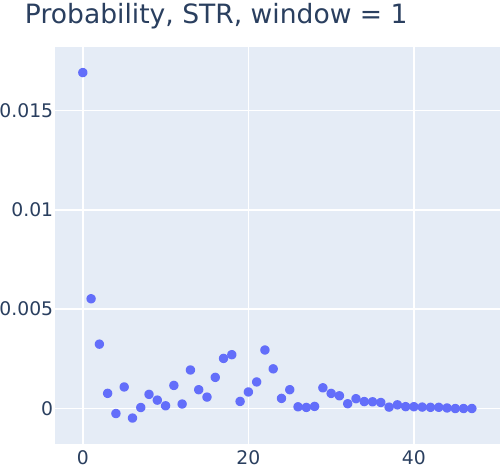}
    \caption{Probability (STR)}
    \end{subfigure}
    \begin{subfigure}[t]{0.23\linewidth}
        \includegraphics[width=\linewidth]{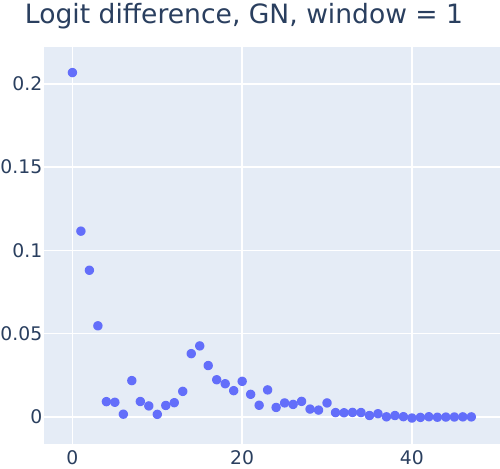}
        \caption{Logit difference (GN)}
    \end{subfigure}
     \begin{subfigure}[t]{0.23\linewidth}
        \includegraphics[width=\linewidth]{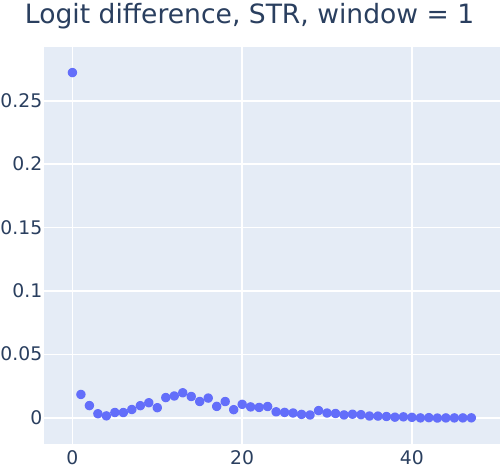}
        \caption{Logit difference (STR)}
    \end{subfigure}
    \caption{\textbf{Patching single MLP layers at the last subject token} in GPT-2 XL on factual recall prompts. None of them suggest a strong peak at the middle MLP layers.}
    \label{fig:single-rome}
\end{figure}

First, we perform single-layer patching of MLP activation at the last subject token and examine the effects in \autoref{fig:single-rome}. We observe that the experiment   suggests  weak or no peak at middle MLP layers, across metrics and corruption methods.

\begin{figure}[ht]
    \centering
    \begin{subfigure}[b]{0.40\linewidth}
    \centering
    \includegraphics[width=\linewidth]{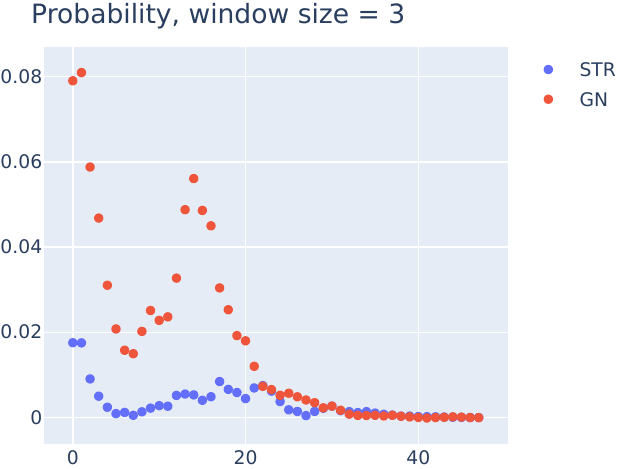}
    \caption{Probability as the metric}
    \end{subfigure}
    \qquad
    \begin{subfigure}[b]{0.40\linewidth}
    \centering
        \includegraphics[width=\linewidth]{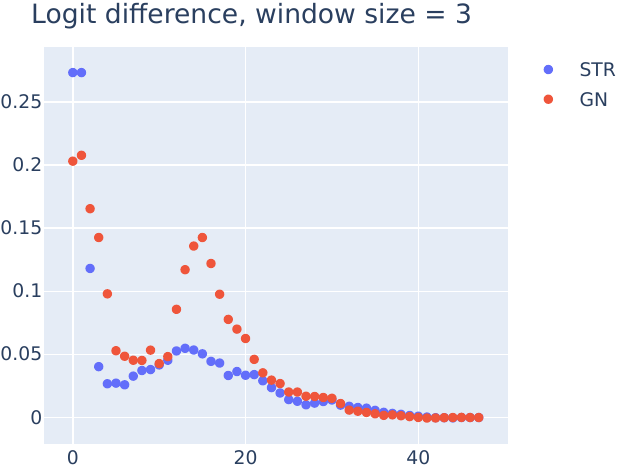}
        \caption{Logit difference as the metric}
    \end{subfigure}
    \caption{\textbf{MLP patching effects  at the  last subject token position} in GPT-2 XL on factual recall prompts, with window size of $3$.}
    \label{fig:mlp-last-3}
\end{figure}

\begin{figure}[ht]
    \centering
    \begin{subfigure}[b]{0.40\linewidth}
    \centering
    \includegraphics[width=\linewidth]{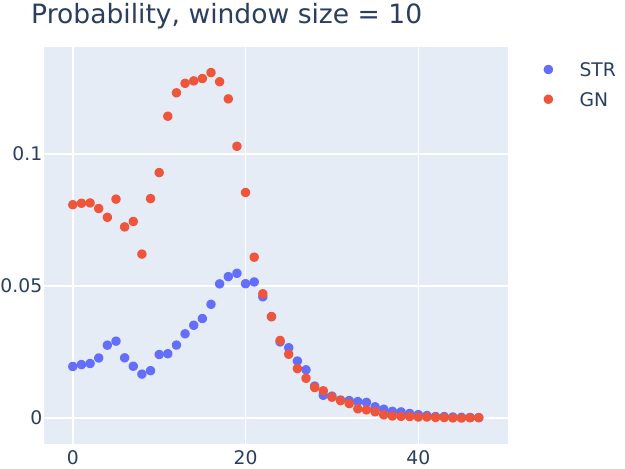}
    \caption{Probability as the metric}
    \end{subfigure}
    \qquad
    \begin{subfigure}[b]{0.40\linewidth}
    \centering
        \includegraphics[width=\linewidth]{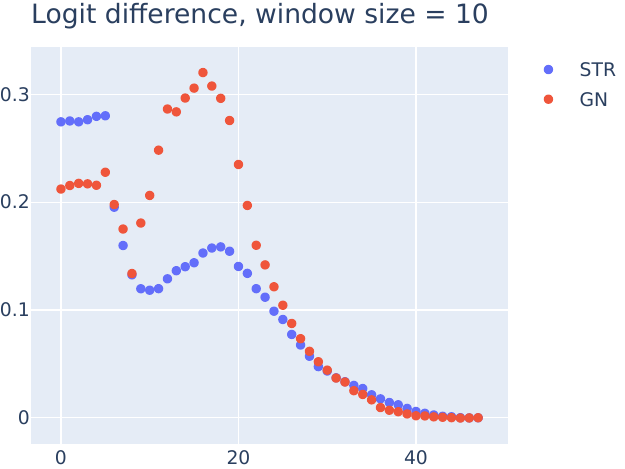}
        \caption{Logit difference as the metric}
    \end{subfigure}
    \caption{\textbf{MLP patching effects for factual recall} at the last subject token position in GPT-2 XL on factual recall prompts, with window size of $10$.}
    \label{fig:mlp-last-10}
\end{figure}

Also, see \autoref{fig:mlp-last-3} and \autoref{fig:mlp-last-10} for plots with sliding window size of $3$ and $10$. Again, activation patching is applied to the MLP activations at the last subject token.
We find again that GN yields significantly more pronounced peak.

%%%%%%%%%%%%%%%%%
%%%%%%%%%%%%%%%%%
\subsection{Plots on MLP patching at all token positions in GPT-2 XL}
\label{sec:b2}
\begin{figure}[ht]
    \centering
    \begin{subfigure}{0.4\linewidth}
    \centering
    \includegraphics[width=\linewidth]{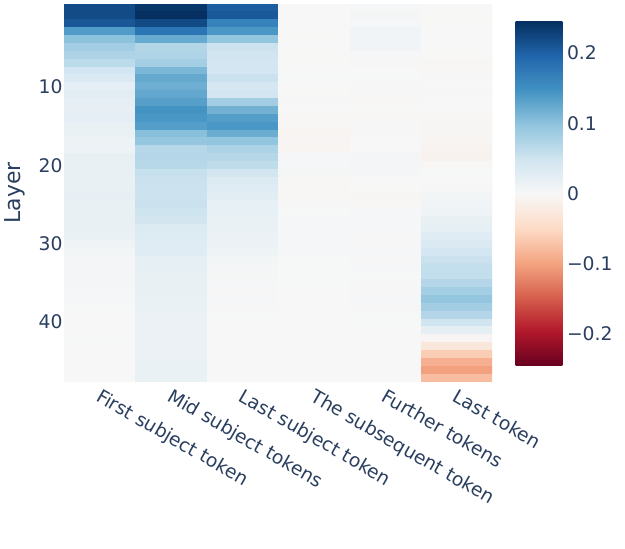}
    \vspace{-2em}
    \caption{Logit difference (GN)}
    \end{subfigure}
    \qquad\qquad
    \begin{subfigure}{0.4\linewidth}
    \centering
        \includegraphics[width=\linewidth]{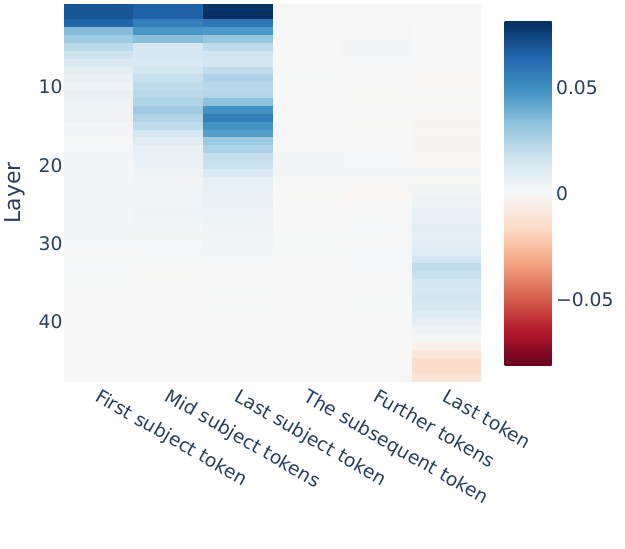}
    \vspace{-2em}
        \caption{Probability (GN)}
    \end{subfigure} 
    \caption{\textbf{Activation patching on MLP} across layers and token positions in GPT-2 XL on factual recall prompts. Apply GN corruption and a sliding window of size $3$.}
    \label{fig:rome-full-gn-3}
\end{figure}

\begin{figure}[ht]
    \centering
    \begin{subfigure}{0.4\linewidth}
    \centering
    \includegraphics[width=\linewidth]{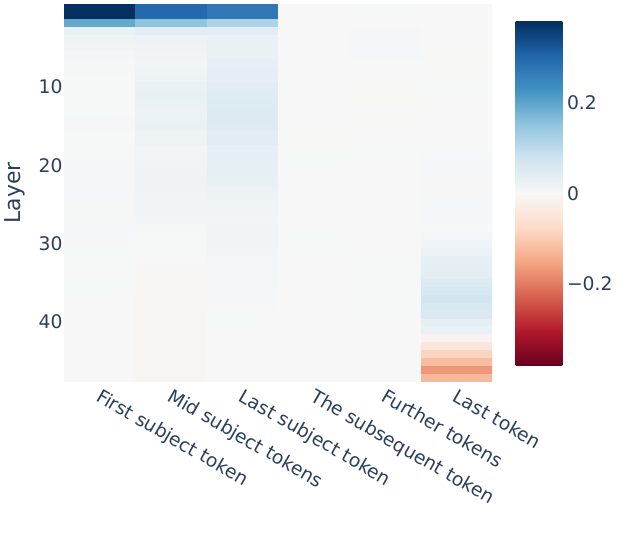}
    \vspace{-2em}
    \caption{Logit difference (STR)}
    \end{subfigure}
    \qquad\qquad
    \begin{subfigure}{0.4\linewidth}
    \centering
        \includegraphics[width=\linewidth]{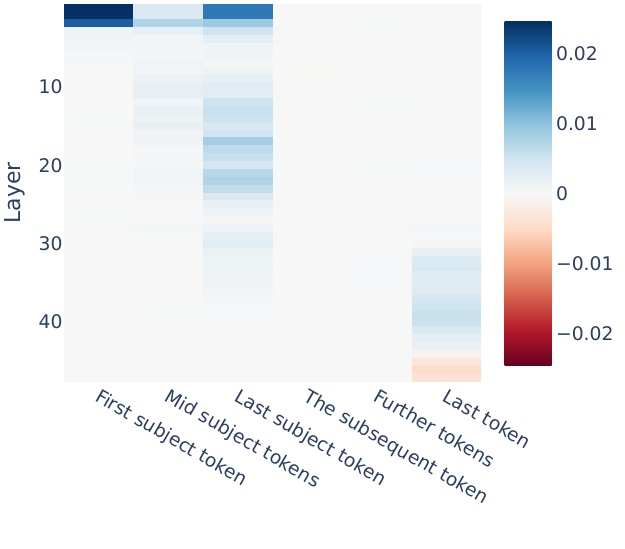}
    \vspace{-2em}
        \caption{Probability (STR)}
    \end{subfigure} 
    \caption{\textbf{Activation patching on MLP} across layers and token positions in GPT-2 XL on factual recall prompts. Apply STR corruption and a sliding window of size $3$.}
    \label{fig:rome-full-str-3}
\end{figure}

\begin{figure}[ht]
    \centering
    \begin{subfigure}{0.4\linewidth}
    \centering
    \includegraphics[width=\linewidth]{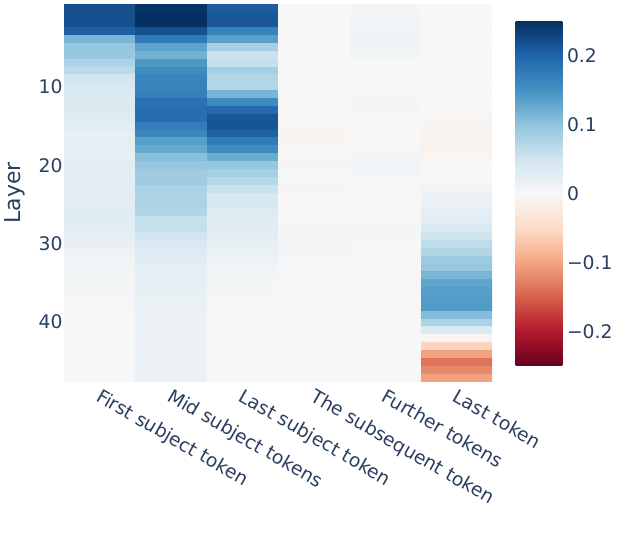}
    \vspace{-2em}
    \caption{Logit difference (GN)}
    \end{subfigure}
    \qquad\qquad
    \begin{subfigure}{0.4\linewidth}
    \centering
        \includegraphics[width=\linewidth]{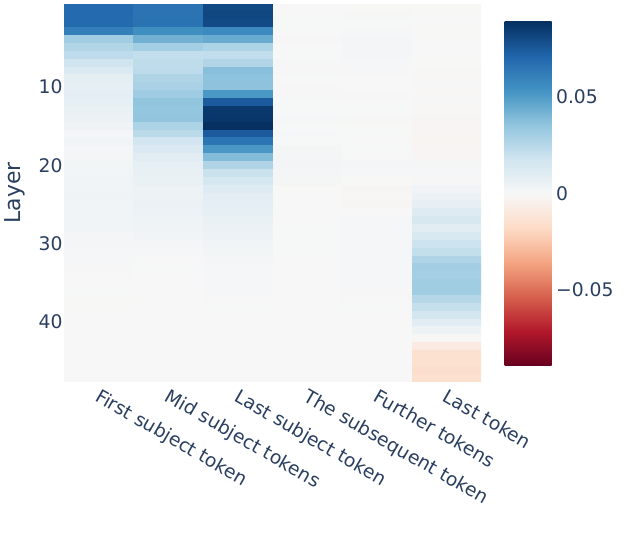}
    \vspace{-2em}
        \caption{Probability (GN)}
    \end{subfigure} 
    \caption{\textbf{Activation patching on MLP} across layers and token positions in GPT-2 XL on factual recall prompts. Apply GN corruption and a sliding window of size $5$.}
    \label{fig:rome-full-gn}
\end{figure}

\begin{figure}[ht]
    \centering
    \begin{subfigure}{0.4\linewidth}
    \centering
    \includegraphics[width=\linewidth]{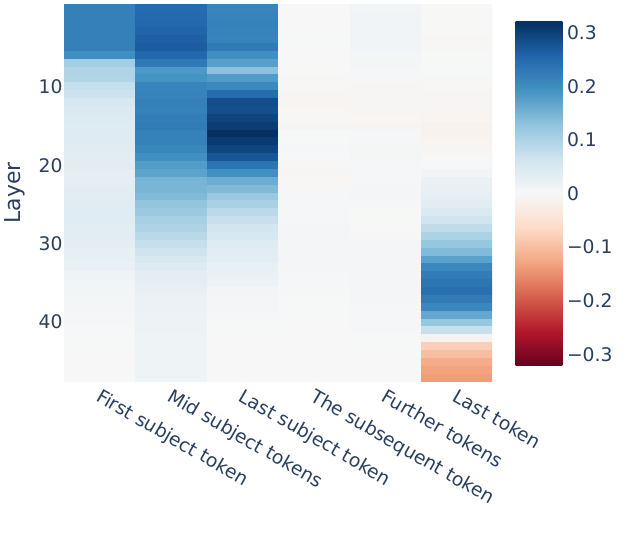}
    \vspace{-2em}
    \caption{Logit difference (GN)}
    \end{subfigure}
    \qquad\qquad
    \begin{subfigure}{0.4\linewidth}
    \centering
        \includegraphics[width=\linewidth]{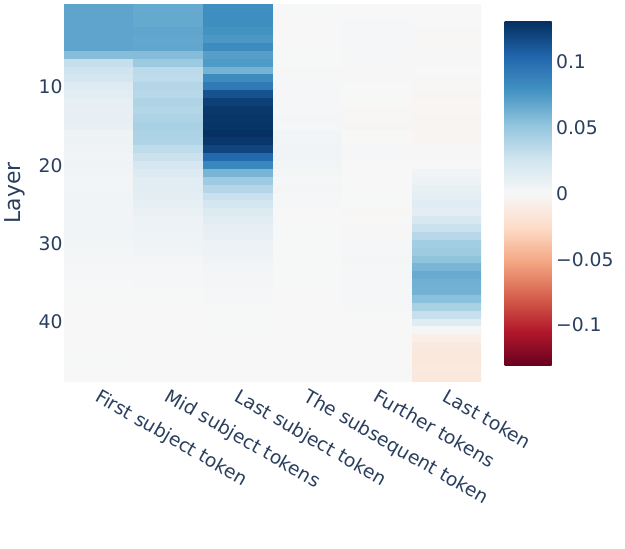}
    \vspace{-2em}
        \caption{Probability (GN)}
    \end{subfigure} 
    \caption{\textbf{Activation patching on MLP} across layers and token positions in GPT-2 XL on factual recall prompts. Apply GN corruption and a sliding window of size $10$.}
    \label{fig:rome-full-gn-10}
\end{figure}

\begin{figure}[ht]
    \centering
    \begin{subfigure}{0.4\linewidth}
    \centering
    \includegraphics[width=\linewidth]{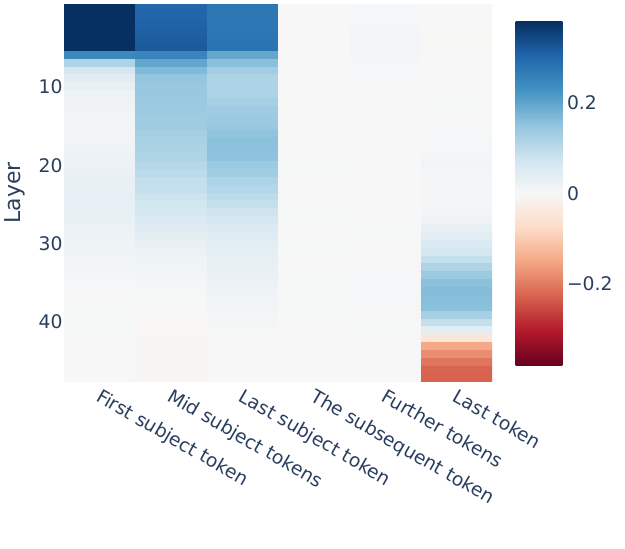}
    \vspace{-2em}
    \caption{Logit difference (STR)}
    \end{subfigure}
    \qquad\qquad
    \begin{subfigure}{0.4\linewidth}
    \centering
        \includegraphics[width=\linewidth]{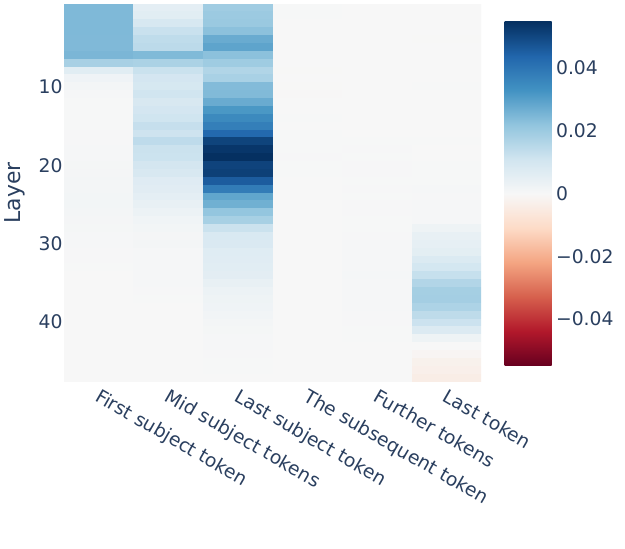}
    \vspace{-2em}
        \caption{Probability (STR)}
    \end{subfigure} 
    \caption{\textbf{Activation patching on MLP} across layers and token positions in GPT-2 XL. Apply STR corruption and a sliding window of size $10$.}
    \label{fig:rome-full-str-10}
\end{figure}

See \autoref{fig:rome-full-gn-3}--\autoref{fig:rome-full-str-10} to plots on MLP patching at all token positions in GPT-2 XL, across window sizes of $3,5,10$. We observe that the right-side plots, using probability as the metric, highlights the last subject token as important. In contrast, the left-side figure using logit different does it to lesser degree.

\subsection{Plots on sliding window patching in GPT2-XL}
\label{sec:sliding-plots}
\begin{figure}[ht]
    \centering
    \begin{subfigure}[b]{0.40\linewidth}
    \centering
    \includegraphics[width=\linewidth]{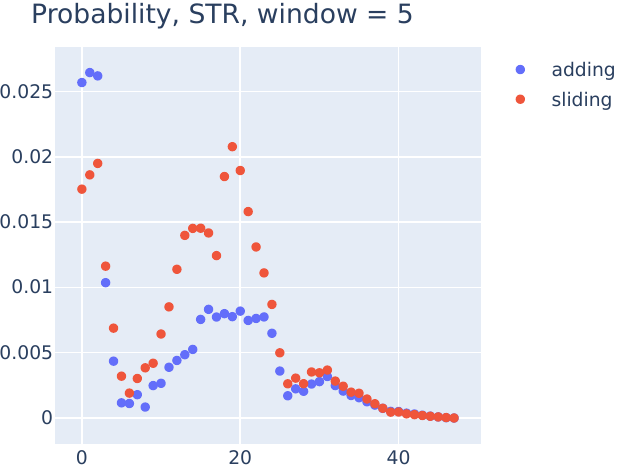}
    \caption{STR corruption}
    \end{subfigure}
    \qquad
    \begin{subfigure}[b]{0.40\linewidth}
    \centering
        \includegraphics[width=\linewidth]{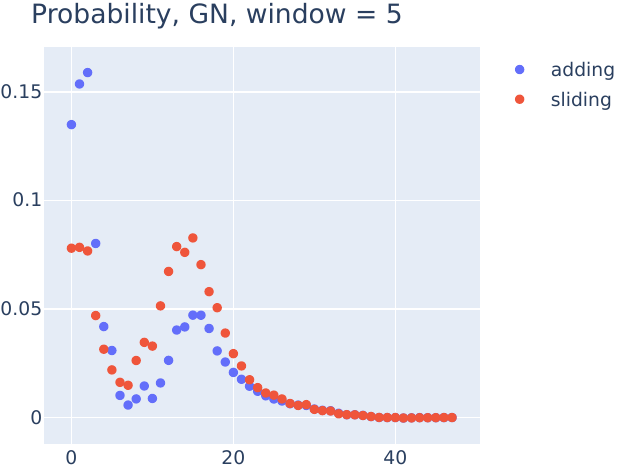}
        \caption{GN corruption}
    \end{subfigure}
    \caption{\textbf{MLP patching effects, sliding window vs summing up single-layer patching} at last token position in GPT-2 XL on factual recall prompts, with window size of $5$. Apply probability as the metric.}
    \label{fig:mlp-last-5-sliding-pr}
\end{figure}

\begin{figure}[ht]
    \centering
    \begin{subfigure}[b]{0.40\linewidth}
    \centering
    \includegraphics[width=\linewidth]{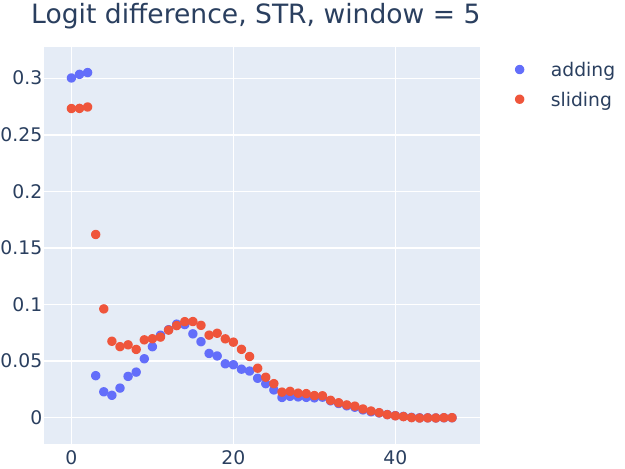}
    \caption{STR corruption}
    \end{subfigure}
    \qquad
    \begin{subfigure}[b]{0.40\linewidth}
    \centering
        \includegraphics[width=\linewidth]{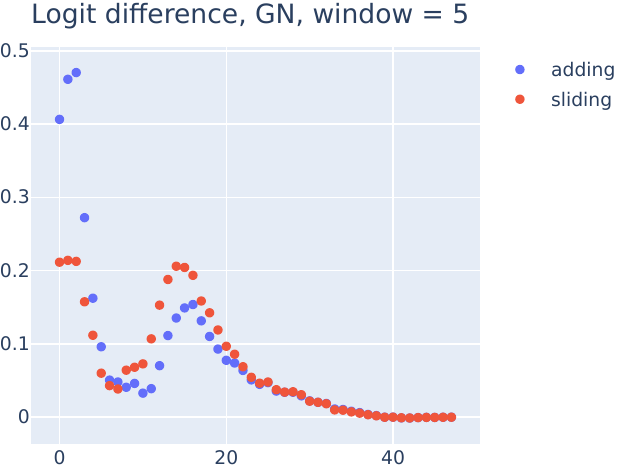}
        \caption{GN corruption}
    \end{subfigure}
    \caption{\textbf{MLP patching effects, sliding window vs summing up single-layer patching} at last token position in GPT-2 XL on factual recall prompts, with window size of $5$. Apply logit difference as the metric.}
    \label{fig:mlp-last-5-sliding-ld}
\end{figure}

\begin{figure}[ht]
    \centering
    \begin{subfigure}[b]{0.40\linewidth}
    \centering
    \includegraphics[width=\linewidth]{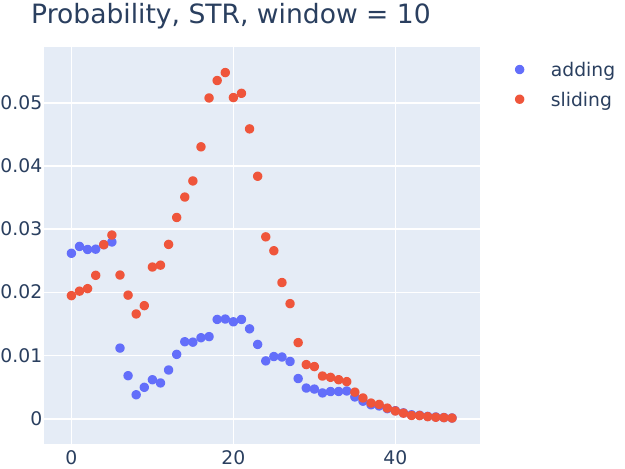}
    \caption{STR corruption}
    \end{subfigure}
    \qquad
    \begin{subfigure}[b]{0.40\linewidth}
    \centering
        \includegraphics[width=\linewidth]{img/rome/sliding-pr-gn-10.pdf}
        \caption{GN corruption}
    \end{subfigure}
    \caption{\textbf{MLP patching effects, sliding window vs summing up single-layer patching} at last token position in GPT-2 XL on factual recall prompts, with window size of $10$. Apply probability as the metric.}
    \label{fig:mlp-last-10-sliding-pr}
\end{figure}

\begin{figure}[ht]
    \centering
    \begin{subfigure}[b]{0.40\linewidth}
    \centering
    \includegraphics[width=\linewidth]{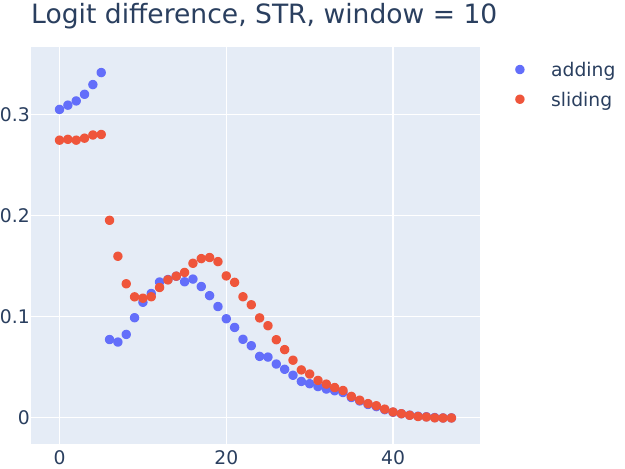}
    \caption{STR corruption}
    \end{subfigure}
    \qquad
    \begin{subfigure}[b]{0.40\linewidth}
    \centering
        \includegraphics[width=\linewidth]{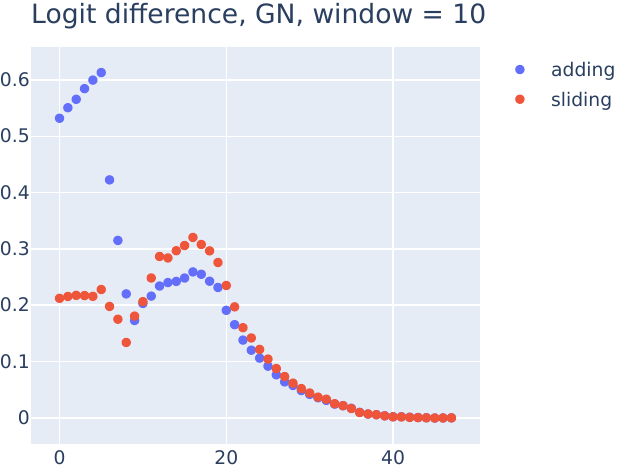}
        \caption{GN corruption}
    \end{subfigure}
    \caption{\textbf{MLP patching effects, sliding window vs summing up single-layer patching} at last token position in GPT-2 XL on factual recall prompts, with window size of $10$. Apply logit difference as the metric.}
    \label{fig:mlp-last-10-sliding-ld}
\end{figure}

We provide further plots from our experiment that compares the sliding window patching
with individual patching aggregated via   summation over windows. See \autoref{fig:mlp-last-5-sliding-pr}--\autoref{fig:mlp-last-10-sliding-ld}.


\subsection{Plots on activation patching of MLP  layers on GPT-2 large}
\label{sec:large-apx}
\begin{figure}[ht]
    \centering
    \begin{subfigure}[b]{0.40\linewidth}
    \centering
    \includegraphics[width=\linewidth]{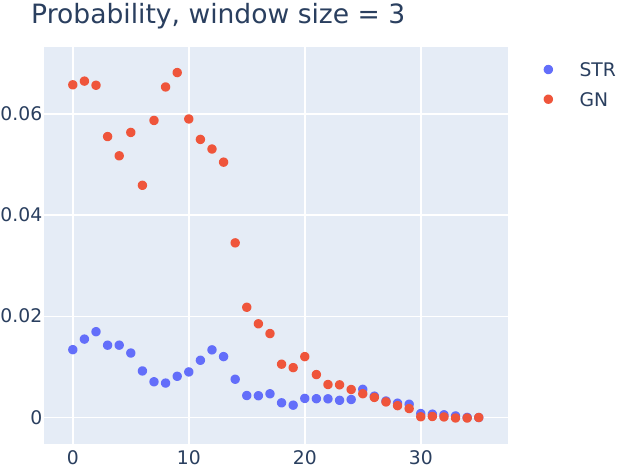}
    \caption{Probability as the metric}
    \end{subfigure}
    \qquad
    \begin{subfigure}[b]{0.40\linewidth}
    \centering
        \includegraphics[width=\linewidth]{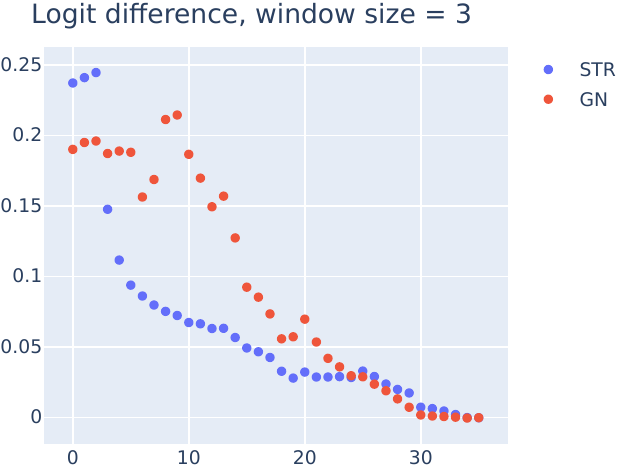}
        \caption{Logit difference as the metric}
    \end{subfigure}
    \caption{\textbf{MLP patching effects  at the last subject token position} in GPT-2 large on factual recall prompts, with window size of $3$.}
    \label{fig:l-3-last}
\end{figure}

\begin{figure}[ht]
    \centering
    \begin{subfigure}[b]{0.40\linewidth}
    \centering
    \includegraphics[width=\linewidth]{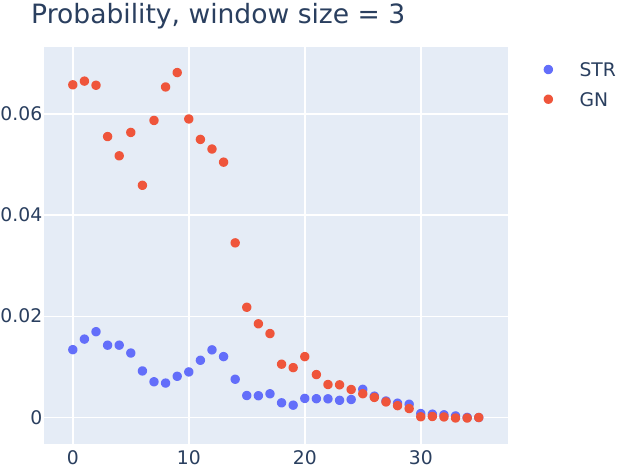}
    \caption{Probability as the metric}
    \end{subfigure}
    \qquad
    \begin{subfigure}[b]{0.40\linewidth}
    \centering
        \includegraphics[width=\linewidth]{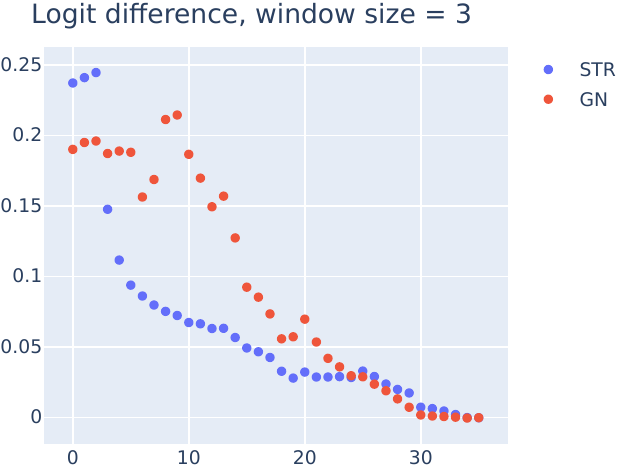}
        \caption{Logit difference as the metric}
    \end{subfigure}
    \caption{\textbf{MLP patching effects  at the last subject token position} in GPT-2 large on factual recall prompts, with window size of $5$.}
    \label{fig:l-5-last}
\end{figure}

We perform activation patching on MLP layers of GPT-2 large   in the factual association setting. 
Following our experiments in \autoref{sec:corruption-diff}, we focus the effects at patching the MLP activation of the last subject token.  
We validate the high-level finding of \autoref{sec:corruption-diff}, where we observe the disparity of GN and STR applied to MLP activation in the  factual prediction setting. In particular, GN gives more pronounced concentration at early-middle MLP layers. 
We apply sliding window patching of size $3$ and $5$; see  \autoref{fig:l-3-last} and \autoref{fig:l-5-last} for the resulting plots.

\subsection{Plots on activation patching of MLP layers in GPT-J}
\label{sec:gpt-j}
\begin{figure}[ht]
    \centering
    \begin{subfigure}{0.4\linewidth}
    \centering
    \includegraphics[width=\linewidth]{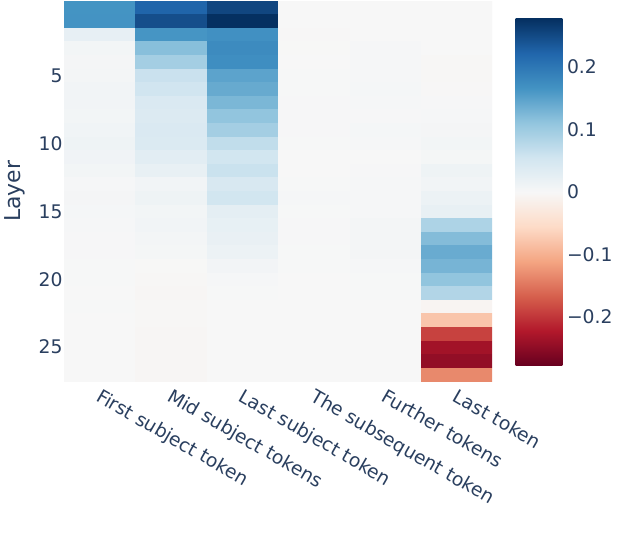}
    \vspace{-2em}
    \caption{Logit difference (GN)}
    \end{subfigure}
    \qquad\qquad
    \begin{subfigure}{0.4\linewidth}
    \centering
        \includegraphics[width=\linewidth]{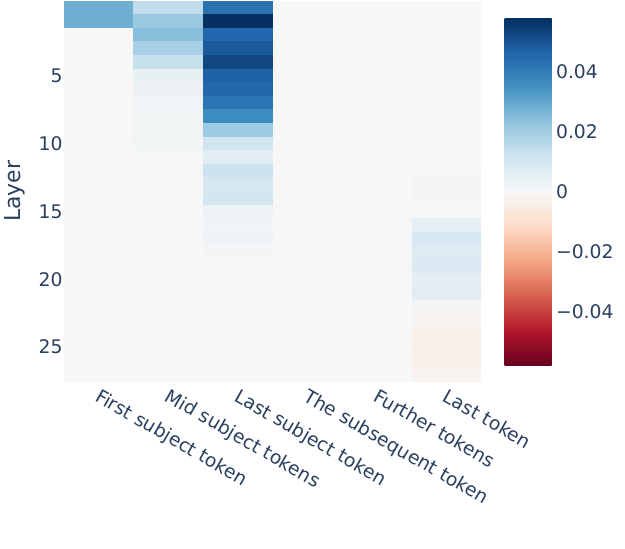}
    \vspace{-2em}
        \caption{Probability (GN)}
    \end{subfigure} 
    \caption{\textbf{Activation patching on MLP across layers and token positions in GPT-J} on factual recall prompts. Apply STR corruption and a sliding window of size $5$.}
    \label{fig:rome-j-str-5}
\end{figure}

\begin{figure}[ht]
    \centering
    \begin{subfigure}{0.4\linewidth}
    \centering
    \includegraphics[width=\linewidth]{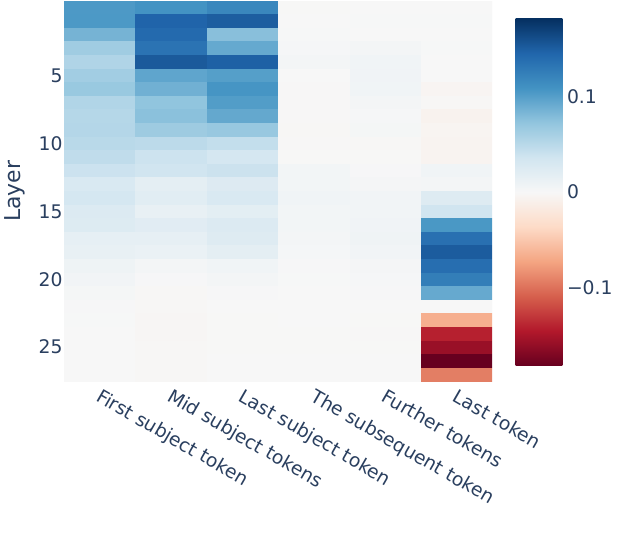}
    \vspace{-2em}
    \caption{Logit difference (GN)}
    \end{subfigure}
    \qquad\qquad
    \begin{subfigure}{0.4\linewidth}
    \centering
        \includegraphics[width=\linewidth]{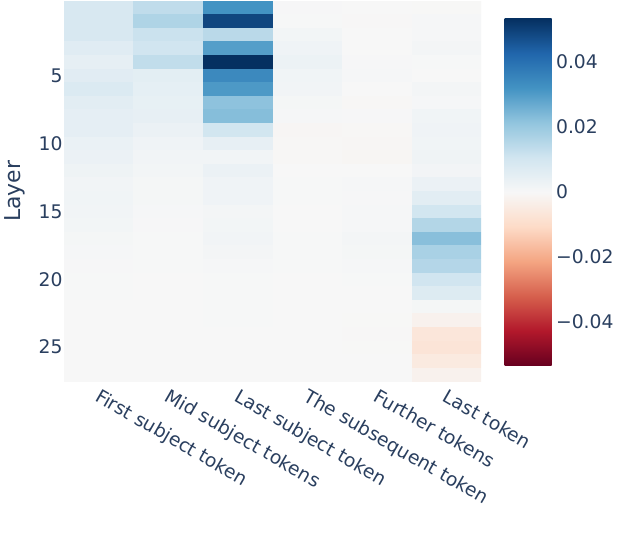}
    \vspace{-2em}
        \caption{Probability (GN)}
    \end{subfigure} 
    \caption{\textbf{Activation patching on MLP across layers and token positions in GPT-J} on factual recall prompts. Apply GN corruption and a sliding window of size $5$.}
    \label{fig:rome-j-gn-5}
\end{figure}

We perform activation patching on MLP layers of GPT-J \citep{gpt-j} in the factual association setting.  We patch the MLP activations across all token positions and verify that probability tends to highlight the importance of the last subject token than logit difference. 
We focus on a sliding window patching of size $5$ and the plots are given in \autoref{fig:rome-j-gn-5} (GN) and \autoref{fig:rome-j-str-5} (STR).  
This complements our results in \autoref{sec:metrics}.

\section{Further details on    IOI circuit discovery}
\label{sec:further-ioi}
\subsection{Detailed plots on activation patching}
\begin{figure}[ht]
    \centering
    \begin{subfigure}{0.32\linewidth}
    \centering
    \includegraphics[width=\linewidth]{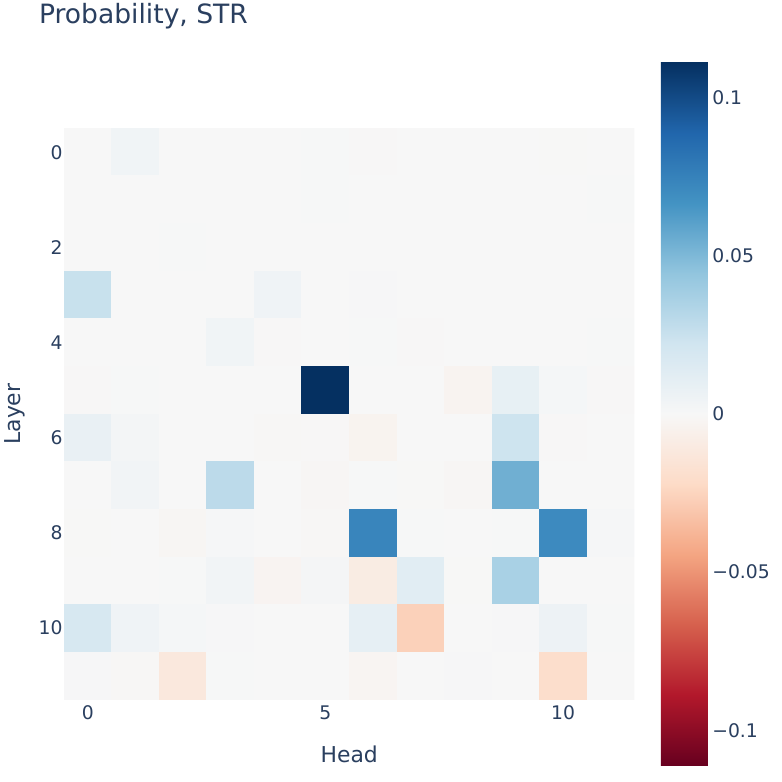}
    \caption{Probability as the metric}
    \end{subfigure}
    \begin{subfigure}{0.32\linewidth}
    \centering
        \includegraphics[width=\linewidth]{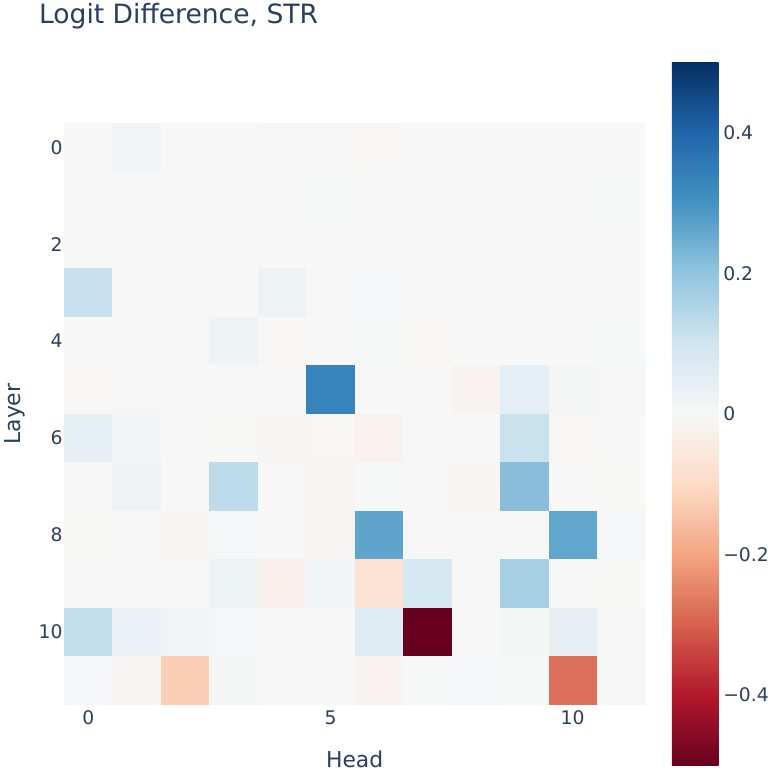}
        \caption{Logit difference as the metric}
    \end{subfigure}
    \begin{subfigure}{0.32\linewidth}
    \centering
        \includegraphics[width=\linewidth]{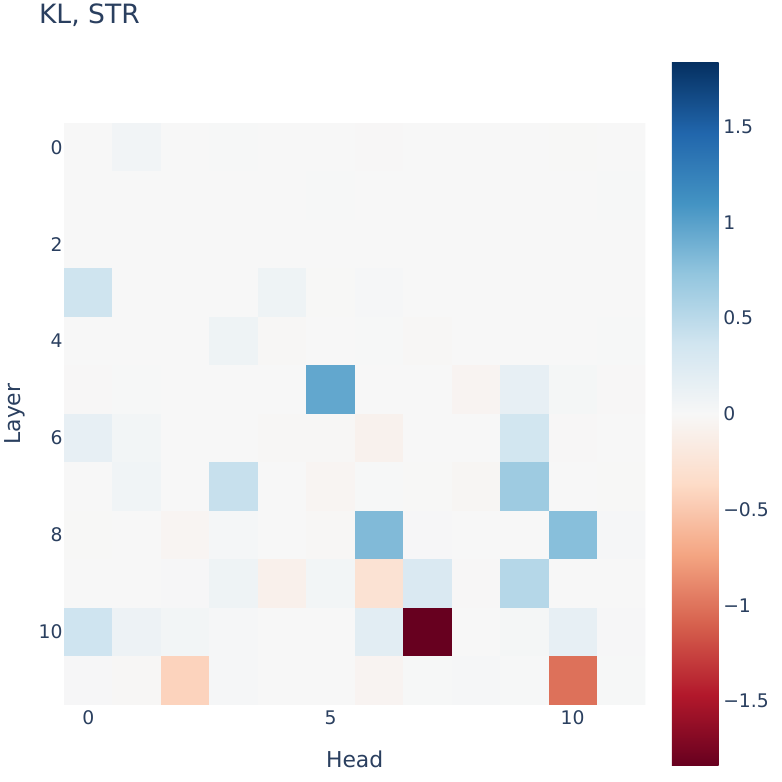}
        \caption{KL divergence as the metric}
    \end{subfigure}
    \caption{\textbf{The effects of patching attention heads} in GPT-2 small using STR corruption on IOI sentences. }
    \label{fig:ioi-str}
\end{figure}

\begin{figure}[ht]
    \centering
    \begin{subfigure}{0.32\linewidth}
    \centering
    \includegraphics[width=\linewidth]{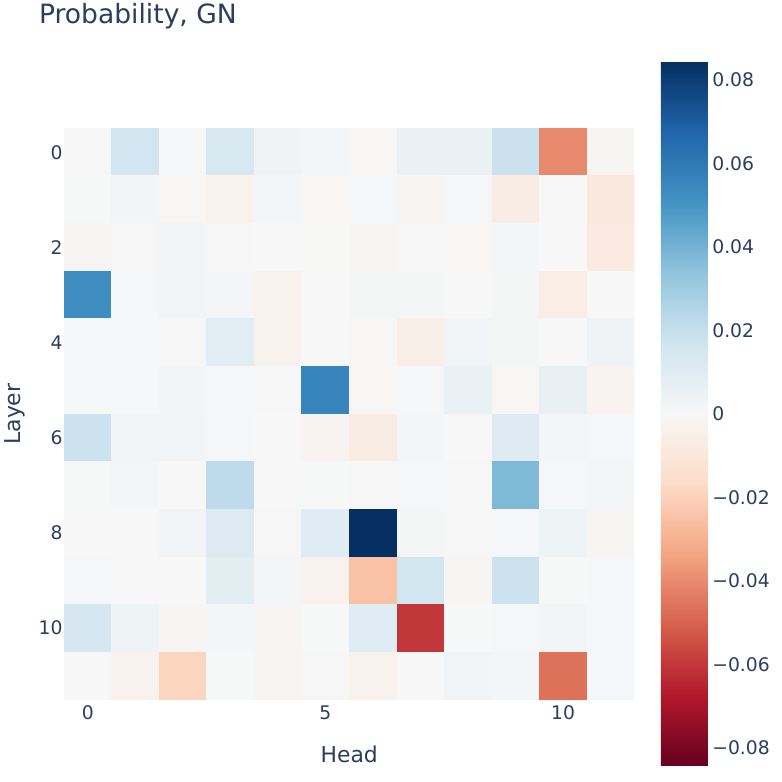}
    \caption{Probability as the metric}
    \end{subfigure}
    \begin{subfigure}{0.32\linewidth}
    \centering
        \includegraphics[width=\linewidth]{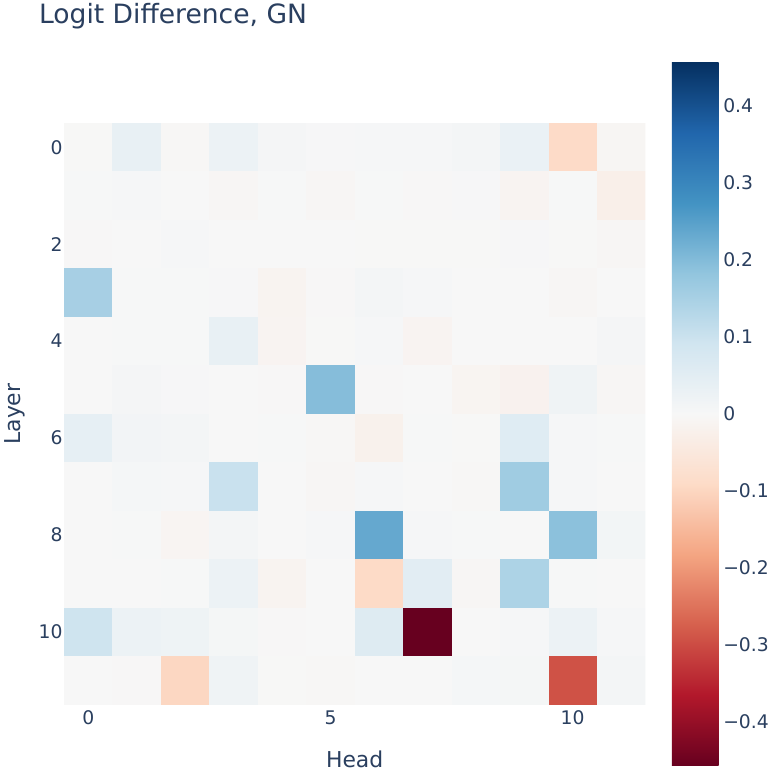}
        \caption{Logit difference as the metric}
        \label{fig:str-010}
    \end{subfigure}
     \begin{subfigure}{0.32\linewidth}
    \centering
        \includegraphics[width=\linewidth]{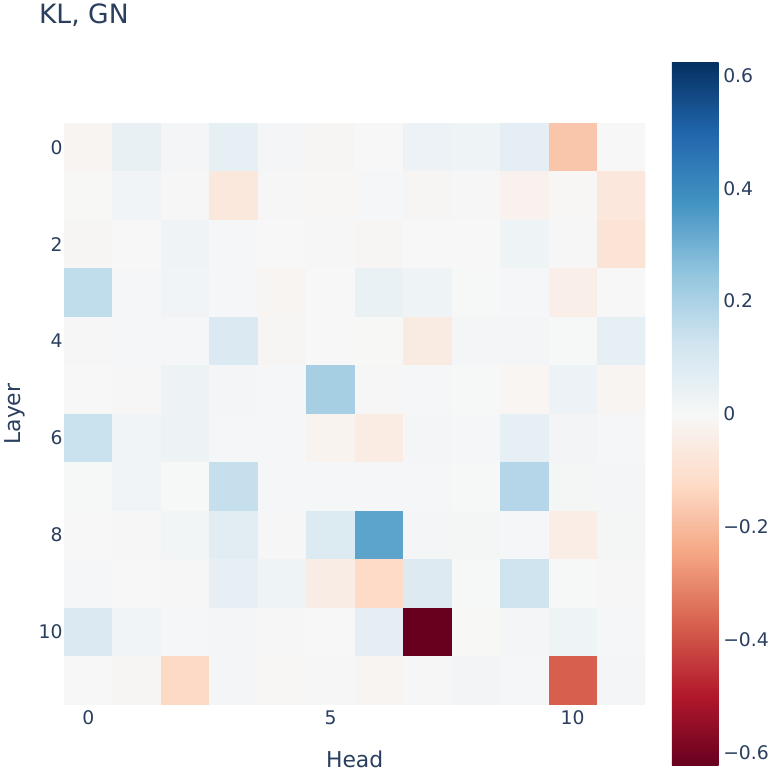}
        \caption{KL divergence as the metric}
    \end{subfigure}
    \caption{\textbf{The effects of patching attention heads} in GPT-2 small using GN corruption on IOI sentences. }
    \label{fig:ioi-gn}
\end{figure}

We now provide the detailed plots from the activation patching experiments on the IOI circuit discovery task \citep{wang2022interpretability}; see \autoref{fig:ioi-str} and \autoref{fig:ioi-gn}.

\subsection{Details on detections}
We provide a detailed list  of detection from attention heads patching in the IOI circuit setting (\autoref{sec:corruption}); see \autoref{tab:ioi-detection-list}.
\label{sec:ioi-detect}
\begin{table}[ht]
\centering
\begin{tabular}{l|l||c|c }
\toprule
\textbf{Corruption}  & \textbf{Metric}      &  \multicolumn{1}{l|}{Negative  heads} &  \multicolumn{1}{l}{Positive heads}    \\ \midrule
STR               & Logit difference          &10.7, 11.10        & 5.5, 7.9, 8.6, 8.10, 9.9                     \\ 
STR               & Probability & 10.7      &5.5, 7.9, 8.6, 8.10, 9.9\\  
STR               & KL divergence          &10.7, 11.10    & 5.5, 7.9, 8.6, 8.10, 9.9    
\\ \midrule
GN                &Logit difference      &     10.7, 11.10       & 3.0, 5.5, 7.9, 8.6, 8.10, 9.9    \\  
GN               & Probability & 0.10, 10.7, 11.10 & 3.0, 5.5, 7.9, 8.6  \\ 
GN               & KL divergence &0.10, 10.7, 11.10     &5.5, 7.9, 8.6 \\ 
\bottomrule
\end{tabular}
\caption{\textbf{Detailed results from attention heads patching} in GPT-2 small on IOI sentences. A head is detected if the patching effect is two standard deviation from the mean effect. Negative heads are heads with negative patching effects, suggesting they hurt model performance.}
\label{tab:ioi-detection-list}
\end{table}

\subsection{Detailed plots on fully random corruption} \label{sec:full-random}
\begin{figure}[ht]
    \centering
    \begin{subfigure}{0.43\linewidth}
    \centering
    \includegraphics[width=\linewidth]{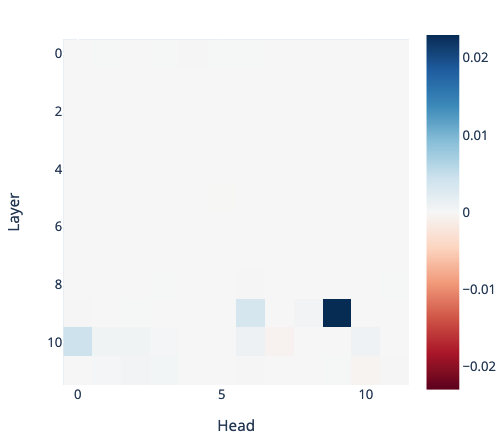}
    \caption{Probability as the metric}
    \end{subfigure}
    \qquad
    \begin{subfigure}{0.43\linewidth}
    \centering
        \includegraphics[width=\linewidth]{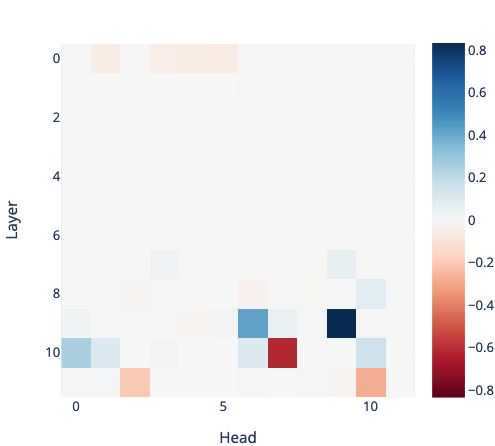}
        \caption{Logit difference as the metric}
    \end{subfigure}
    \caption{\textbf{The effects of patching attention heads}  in GPT-2 small using fully random  corruption on IOI sentences, with S1, S2 and IO replaced by three random names (denoted by $p_{\text{ABC}}$ in \cite{wang2022interpretability}).}
    \label{fig:ioi-full-random}
\end{figure}

We provide the plots on fully random corruption, termed $p_\text{ABC}$ in \cite{wang2022interpretability}. We perform activation patching on all attention heads, using both probability and logit difference as the metric in order to draw contrasts between them.  See \autoref{fig:ioi-full-random}. In particular, we notice that there is no negative head in the plot. This is natural and totally expected, as we explained in \autoref{sec:metrics}.

\section{Dataset Samples}
\label{sec:dataset}

\paragraph{Factual data}
We list a few dataset examples from the \textsc{PairedFacts} dataset used in the factual recall experiments in \autoref{fig:sample-pairs}.\footnote{The full dataset is available at {\small\url{https://www.jsonkeeper.com/b/P1GL}}.} All the prompts are known true facts.
\begin{figure}[htbp]
\begin{subfigure}[t]{0.5\linewidth}
 \begin{lstlisting} 
{
  "pair": [
    "Honus Wagner professionally plays the sport of",
    "Don Shula professionally plays the sport of"
  ],
  "answer": [
    " baseball",
    " football"
  ],
  "length": 9,
  "category": "athletes"
} 
 \end{lstlisting}       
    \end{subfigure}
    \hspace{0.5em}
    \begin{subfigure}[t]{0.5\linewidth}
 \begin{lstlisting}
{
  "pair": [
    "Wii MotionPlus is developed by",
    "Chromebook Pixel is developed by"
  ],
  "answer": [
    " Nintendo",
    " Google"
  ],
  "length": 8,
  "category": "developers"
}
 \end{lstlisting}
\end{subfigure}
    \begin{subfigure}[t]{0.5\linewidth}
        \begin{lstlisting}
{
  "pair": [
    "Schreckhorn belongs to the continent of",
    "Afghanistan belongs to the continent of"
  ],
  "answer": [
    " Europe",
    " Asia"
  ],
  "length": 9,
  "category": "continents"
} 
\end{lstlisting}
    \end{subfigure}
    \hspace{0.5em}
\begin{subfigure}[t]{0.5\linewidth}
\begin{lstlisting}
{
  "pair": [
    "The Eiffel Tower is in the city of",
    "Kinkakuji Temple is in the city of"
  ],
  "answer": [
    " Paris",
    " Kyoto"
  ],
  "category": "city_landmarks",
  "length": 11
}
\end{lstlisting}
    \end{subfigure}
    \caption{\textbf{Sample text prompts} from the \textsc{PairedFacts} dataset. The {\small\texttt{length}} field refers to the sequence length of the prompt under GPT-2 tokenizer.}
    \label{fig:sample-pairs}
\end{figure}

\paragraph{IOI circuit}
The detailed templates of constructing the $p_{\text{IOI}}$ data distribution can be found in Appendix A of \cite{wang2022interpretability}. We perform the same procedure of generating the IOI data   by simply reusing their original code.
\end{document}